\documentclass[10pt,twocolumn,letterpaper]{article}

\usepackage{wacv}
\usepackage{times}
\usepackage{epsfig}
\usepackage{graphicx}
\usepackage{amsmath}
\usepackage{amssymb}
\usepackage{pifont}
\usepackage{rotating}
\usepackage{array}
\usepackage{nicefrac}
\usepackage{xcolor}
\usepackage[percent]{overpic}

\usepackage[accsupp]{axessibility}  % Improves PDF readability for those with disabilities.

\def\wacvPaperID{770} % Enter the WACV Paper ID here

\wacvfinalcopy % *** Uncomment this line for the final submission

\ifwacvfinal
\fi

\ifwacvfinal
\usepackage[breaklinks=true,bookmarks=false]{hyperref}
\else
\usepackage[pagebackref=true,breaklinks=true,colorlinks,bookmarks=false]{hyperref}
\fi

\ifwacvfinal
\usepackage[breaklinks=true,bookmarks=false]{hyperref}
\else
\usepackage[pagebackref=true,breaklinks=true,colorlinks,bookmarks=false]{hyperref}
\fi

\begin{document}

%%%%%%%%% TITLE
\title{SeaDronesSee: A Maritime Benchmark for Detecting Humans in Open Water}

\author{{Leon Amadeus Varga\thanks{These authors contributed equally to this work. The order of names is determined by coin flipping}, Benjamin Kiefer\footnotemark[1], Martin Messmer\footnotemark[1], Andreas Zell}\\
\textit{Cognitive Systems Group}\\
\textit{University of Tuebingen} \\
Tuebingen, Germany\\
Email: {\tt\small leon.varga@uni-tuebingen.de}, {\tt\small benjamin.kiefer@uni-tuebingen.de},\\ {\tt\small martin.messmer@uni-tuebingen.de}, {\tt\small andreas.zell@uni-tuebingen.de}
}

\maketitle
\thispagestyle{empty}

%%%%%%%%% ABSTRACT
\begin{abstract}
   Unmanned Aerial Vehicles (UAVs) are of crucial importance in search and rescue missions in maritime environments due to their flexible and fast operation capabilities. Modern computer vision algorithms are of great interest in aiding such missions.
   However, they are dependent on large amounts of real-case training data from UAVs, which is only available for traffic scenarios on land. Moreover, current object detection and tracking data sets only provide limited environmental information or none at all, neglecting a valuable source of information. Therefore, this paper introduces a large-scaled visual object detection and tracking benchmark (SeaDronesSee) aiming to bridge the gap from land-based vision systems to sea-based ones. We collect and annotate over 54,000 frames with 400,000 instances captured from various altitudes and viewing angles ranging from 5 to 260 meters and 0 to 90$^\circ$ degrees while providing the respective meta information for altitude, viewing angle and other meta data. 
   We evaluate multiple state-of-the-art computer vision algorithms on this newly established benchmark serving as baselines. We provide an evaluation server where researchers can upload their prediction and compare their results on a central leaderboard \footnote{The leaderboard, the data set and the code to reproduce our results are available at \url{https://seadronessee.cs.uni-tuebingen.de}.
   }.

   %-which tasks-noframes...\url{https://bit.ly/3eNe1NY}
\end{abstract}

%%%%%%%%% BODY TEXT
\section{Introduction}

Unmanned Aerial Vehicles (UAVs) equipped with cameras have grown into an important asset in a wide range of fields, such as agriculture, delivery, surveillance, and search and rescue (SAR) missions \cite{adao2017hyperspectral,san2018uav,geraldes2019uav}. In particular, UAVs are capable of assisting in SAR missions due to their fast and versatile applicability while providing an overview over the scene \cite{mishra2020drone,karaca2018potential,albanese2020sardo}. Especially in maritime scenarios, where wide areas need to be quickly overseen and searched, the efficient use of autonomous UAVs is crucial \cite{yeong2015review}. Among the most challenging issues in this application scenario is the detection, localization, and tracking of people in open water \cite{gallego2019detection,nasr2019shipwrecked}. The small size of people relative to search radii and the variability in viewing angles and altitudes require robust vision-based systems. 

\begin{figure}[t]
	
	\centering
	% left, bottom, right, top
%{\includegraphics[scale=0.1,trim=0 100 0 0,clip]{DSC00407.JPG}}
{\includegraphics[scale=0.1,trim=100 290 0 0,clip]{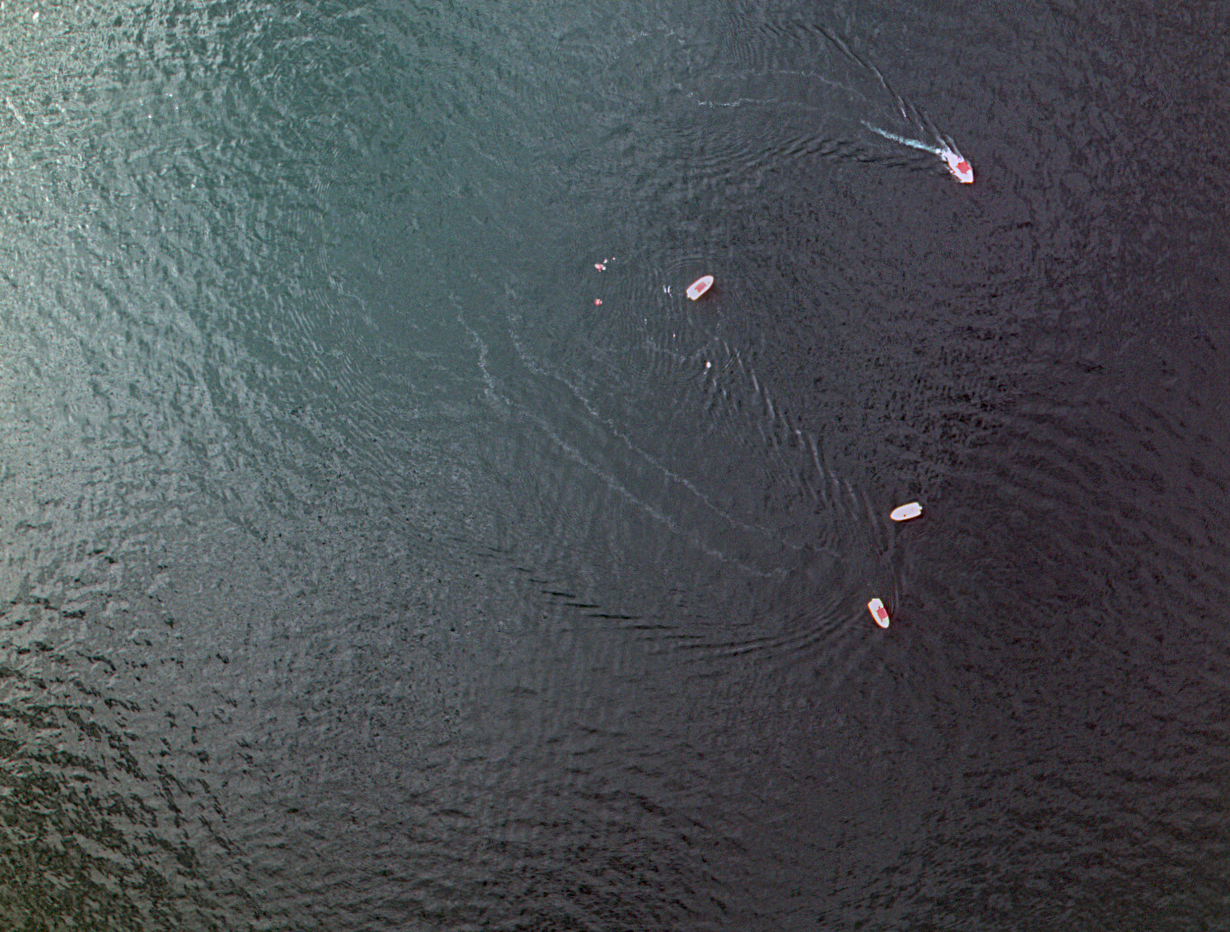}}
{\includegraphics[scale=0.12]{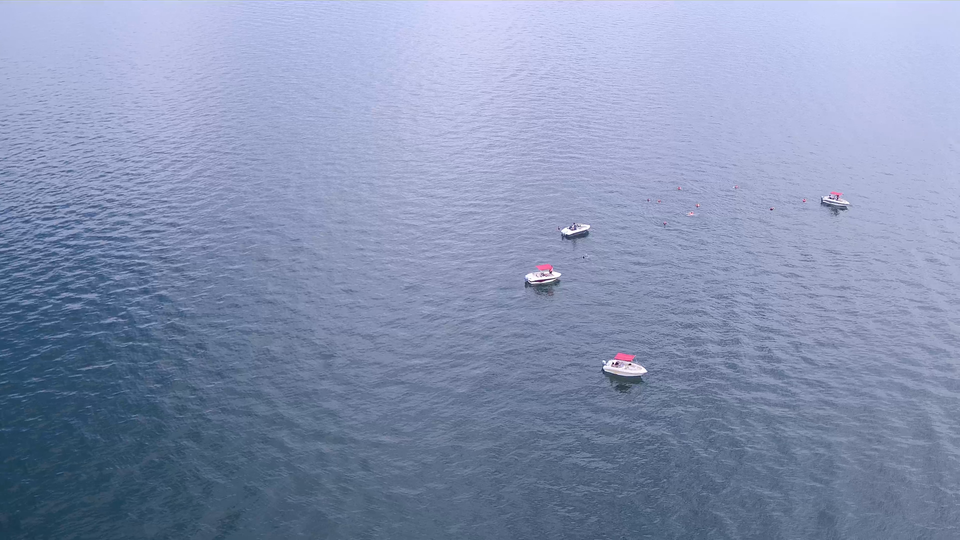}}\\ 
\vspace{0.5mm}
{\includegraphics[scale=0.12,trim=22 0 0 0,clip]{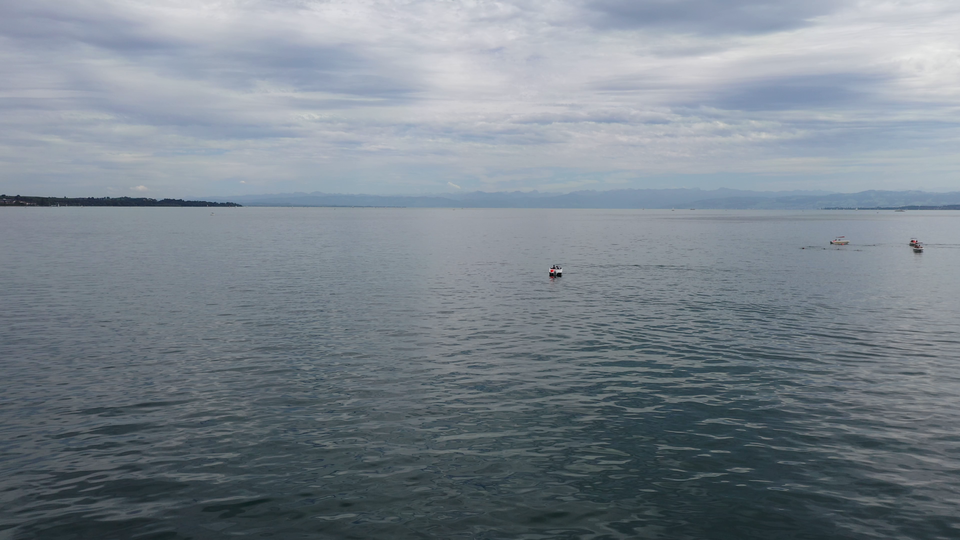}}
{\includegraphics[scale=0.12]{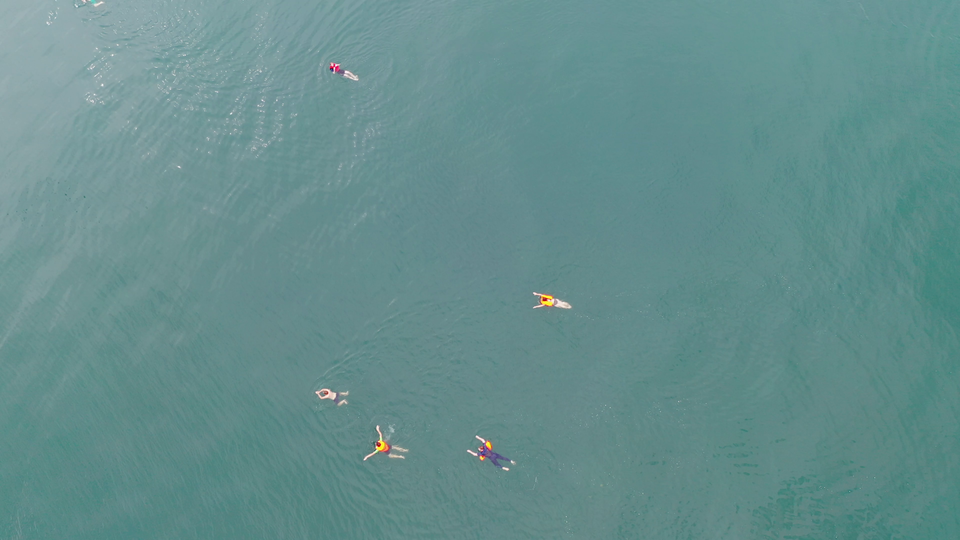}}\\
(a)

{\includegraphics[scale=0.09,trim=21 200 0 0,clip]{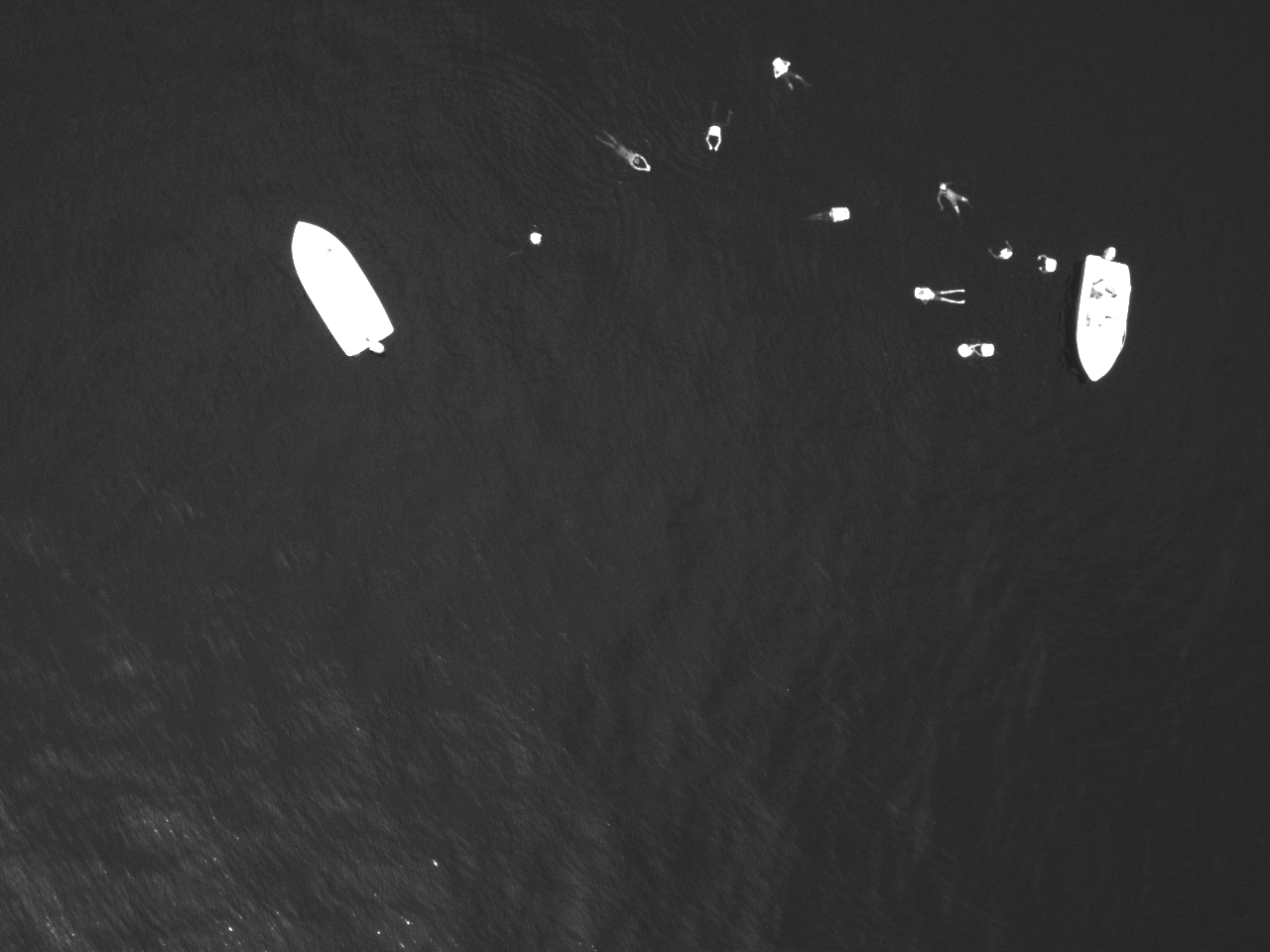}}
{\includegraphics[scale=0.09,trim=0 200 0 0,clip]{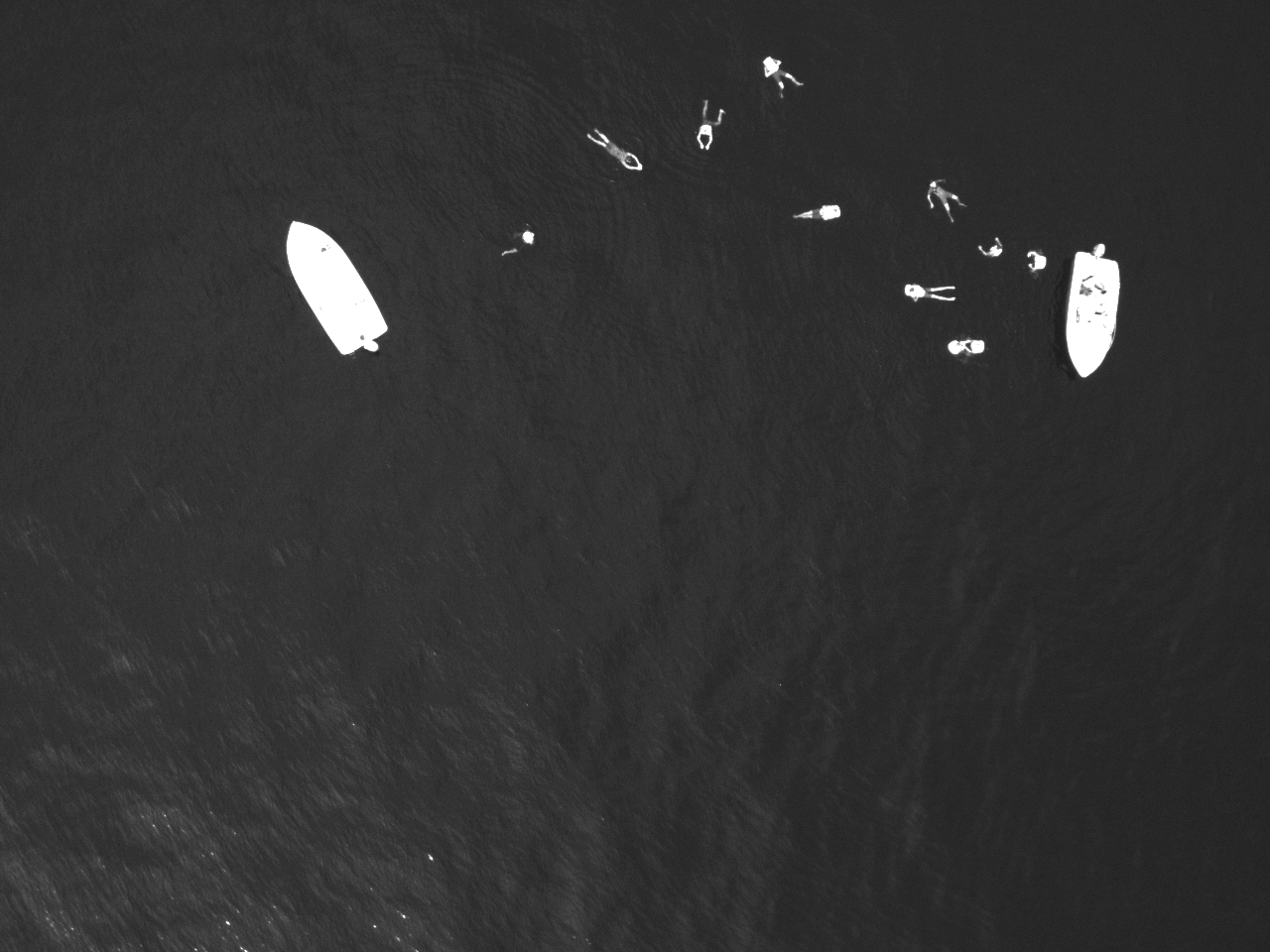}}\\
(b)

	\caption{(a) Typical image examples with varying altitudes and angles of view: 250 m, $90^\circ$; 50 m, $30^\circ$; 10 m, $0^\circ$ and 20 m, $90^\circ$ (from top left to bottom right). (b) Examples of the Red Edge (717 nm, left) and Near Infrared (842 nm, right) light spectra of an image captured by the MicaSense RedEdge-MX. Note the glowing appearance of the swimmers.}
	\label{fig:front_image}
	\vspace{-0.4cm}
\end{figure}

Currently, these systems are implemented via data-driven methods such as deep neural networks. These methods depend on large-scale data sets portraying real-case scenarios to obtain realistic imagery statistics. However, there is a great lack of large-scale data sets in maritime environments. Most data sets captured from UAVs are land-based, often focusing on traffic environments, such as VisDrone \cite{zhu2018vision} and UAVDT \cite{du2018unmanned}. Many of the few data sets that are captured in maritime environments fall in the category of remote sensing, often leveraging satellite-based synthetic aperture radar \cite{crisp2004state}. All of these are only valuable for ship detection \cite{corbane2010complete} as they don't provide the resolution needed for SAR missions. Furthermore, satellite-based imagery is susceptible to clouds and only provides top-down views. Finally, many current approaches in the maritime setting rely on classical machine learning methods, incapable of dealing with the large number of influencing variables and calling for more elaborate models \cite{prasad2019object}.

This work aims to close the gap between large-scale land-based data sets captured from UAVs to maritime-based data sets. We introduce a large-scale data set of people in open water, called SeaDronesSee. We captured videos and images of swimming probands in open water with various UAVs and cameras. As it is especially critical in SAR missions to detect and track objects from a large distance, we captured the RGB footage with 3840$\times$2160 px to 5456$\times$3632 px resolution. We carefully annotated ground-truth bounding box labels for objects of interest including swimmer, floater (swimmer with life jacket), life jacket, swimmer$^\dagger$ (person on boat not wearing a life jacket), floater$^\dagger$ (person on boat wearing a life jacket), and boat.

Moreover, we note that current data sets captured from UAVs only provide very coarse or no meta information at all. We argue that this is a major impediment in the development of multi-modal systems, which take these additional information into account to improve accuracy or speed. Recently, methods that rely on these meta data were proposed. However, they note the lack of large-scaled publicly available data set in that regime  (see \eg \cite{kiefer2021leveraging,wu2019delving,messmer2021gaining}). Therefore, we provide precise meta information for every frame and image including altitude, camera angle, speed, time, and others.

In maritime settings, the use of multi-spectral cameras with Near Infrared channels to detect humans can be advantageous \cite{gallego2019detection}. For that reason, we also captured multi-spectral images using a MicaSense RedEdge. This enables the development of detectors taking into account the non-visible light spectra Near Infrared (842 nm) and Red Edge (717 nm).

Finally, we provide detailed statistics of the data set and conduct extensive experiments using state-of-the-art models and hereby establish baseline models. These serve as a starting point for our SeaDronesSee benchmark. We release the training and validation sets with complete bounding box ground truth but only the test set's videos/images. The ground truth of the test set is used by the benchmark server to calculate the generalization power of the models. We set up an evaluation web page, where researchers can upload their predictions and opt to publish their results on a central leader board such that transparent comparisons are possible. The benchmark focuses on three tasks: (i) object detection, (ii) single-object tracking and (iii) multi-object tracking, which will be explained in more detail in the subsequent sections. Our main contributions are as follows:
\begin{itemize}
	\item To the best of our knowledge, SeaDronesSee is the first large annotated UAV-based data set of swimmers in open water. It can be used to further develop detectors and trackers for SAR missions.
	
	\item We provide full environmental meta information for every frame making SeaDroneSee the first UAV-based data set of that nature.
	
	\item We provide an evaluation server to prevent researches from overfitting and allow for fair comparisons.
	
	\item We perform extensive experiments on state-of-the-art object detectors and trackers on our data set.
	
\end{itemize}

%-------------------------------------------------------------------------

\begin{table*}
	
	\begin{center}
		
		\begin{tabular}{c|c|c|c|cc|cc|c}
			Object detection  & Env. &  Platform  & Image widths   & Altitude & Range & Angle & Range &  Other meta\\
			\hline
			DOTA \cite{xia2018dota} & cities & satellite & 800-20,000 & -- & -- & \ding{53} & $90^\circ$ & \ding{53} \\
				%spatial resolution for dota available anscheiend
			UAVDT \cite{du2018unmanned} & traffic & UAV & 1,024 & \ding{53} & 5-200 m* & \ding{53} & $0-90^\circ$* & \ding{53}\\
			%coarse meta data 
			VisDrone \cite{zhu2018vision} & traffic & UAV & 960-2,000 & \ding{53} & 5-200 m* & \ding{53} & $0-90^\circ$* & \ding{53}\\
			
			Airbus Ship \cite{airbus-ship} & maritime & satellite & 768 & -- & -- & \ding{53} & $90^\circ$\hphantom{*} & \ding{53}\\
			
			AU-AIR \cite{bozcan2020air} & traffic & UAV & 1,920 & \checkmark & 5-30 m\hphantom{*} & \ding{53} & $45-90^\circ$\hphantom{*} & \checkmark \\
			
			\bf SeaDronesSee & maritime & UAV  & 3,840-5,456 & \checkmark & 5-260 m\hphantom{*} & \checkmark & $\ 0-90^\circ$\hphantom{*} & \checkmark  \\				
		\end{tabular}
	\end{center}
	
	\begin{center}
		
		\begin{tabular}{c|c|c|c|cc|cc|c}
			Single-object tracking  & Env.  & \#Clips
			%& \#seq. & \#frames 
			& Frame widths   & Altitude & Range & Angle & Range & Other meta\\
			\hline
			
			UAV123 \cite{mueller2016benchmark} & traffic & 123 &
			%123 & 110 k &
			1,280 & \ding{53} & 5-50 m* & \ding{53} &  $0-90^\circ$* & \checkmark \\
			
			DTB70 \cite{li2017visual} & sports & 70 & 1,280 & \ding{53} & 0-10 m* & \ding{53} & $0-90^\circ$* & \ding{53}\\
			
			UAVDT-SOT \cite{du2018unmanned} & traffic & 50 &
			%50 & 37 k &
			1,024 & \ding{53} & 5-200 m* & \ding{53} &  $0-90^\circ$* & \checkmark \\
			
			VisDrone \cite{zhu2018vision} & traffic & 167 &
			%167 & 139 k &
			960-2,000   &\ding{53} & 5-200 m* & \ding{53} & $0-90^\circ$* & \checkmark  \\
			
			\bf SeaDronesSee & maritime & 208 &
			%274 & 29 k &
			3,840 & \checkmark & 5-150 m\hphantom{*} & \checkmark & $0-90^\circ$\hphantom{*} & \checkmark  \\

		\end{tabular}
	\end{center}
	
	\begin{center}
		
		\begin{tabular}{c|c|c|c|cc|cc|c}
			Multi-object tracking  & Env. & \#Frames
			%& \#seq. & \#frames 
			& Frame widths   & Altitude & Range & Angle & Range & Other meta\\
			\hline
			
			UAVDT-MOT \cite{du2018unmanned} & traffic & 40.7 k &
			%50 & 37 k &
			1,024 & \ding{53} & 5-200 m* & \ding{53} &  $0-90^\circ$* & \checkmark \\
			
			VisDrone \cite{zhu2018vision} & traffic & 40 k &
			%167 & 139 k &
			960-2,000   &\ding{53} & 5-200 m* & \ding{53} & $0-90^\circ$* & \checkmark  \\
			
			\bf SeaDronesSee & maritime & 54 k &
			%274 & 29 k &
			3,840 & \checkmark & 5-150 m\hphantom{*} & \checkmark & $0-90^\circ$\hphantom{*} & \checkmark  \\

		\end{tabular}
	\end{center}

	\caption{Comparison with the most prominent annotated aerial data sets. 'Altitude' and 'Angle' indicate whether or not there are precise altitude and angle view information available. 'Other meta' refers to time stamps, GPS, and IMU data and in the case of object tracking can also mean attribute information about the sequences. The values with stars have been estimated based on ground truth bounding box sizes and corresponding real world object sizes (for altitude) and qualitative estimation of sample images (for angle). For DOTA and Airbus Ship the range of altitudes is not available because these are satellite-based data sets.}
	
	\label{table:comparison_datasets}
\end{table*}

\section{Related Work}
In this section, we review major labeled data sets in the field of computer vision from UAVs and in maritime scenarios which are usable for supervised learning models.

\subsection{Labeled Data Sets Captured from UAVs}
Over the last few years, quite a few data sets captured from UAVs have been published. The most prominent are these that depict traffic situations, such as VisDrone \cite{zhu2018vision} and UAVDT \cite{du2018unmanned}. Both data sets focus on object detection and object tracking in unconstrained environments. Pei \etal \cite{pei2019human}  collect videos (Stanford Drone Dataset) showing traffic participants on campuses (mostly people) for human trajectory prediction usable for object detection. UAV123 \cite{mueller2016benchmark} is a single-object tracking data set consisting of 123 video sequences with corresponding labels. The clips mainly show traffic scenarios and common objects. Both, Hsieh \etal \cite{hsieh2017drone} and Mundhenk \etal \cite{mundhenk2016large} capture a data set showing parking lots for car counting tasks and constrained object detection. Li \etal \cite{li2017visual} provide a single-object tracking data set showing traffic, wild life and sports scenarios. 
%Flying objects (such as footballs) are shown in \cite{rozantsev2016detecting} also for single-object tracking. 
Collins \etal capture a single-object tracking data set showing vehicles on streets in rural areas. Krajewski \etal \cite{krajewski2018highd} show vehicles on freeways.

Another active area of research focuses on drone-based wildlife detection. Van \etal \cite{van2014nature} release a data set for the tasks of low-altitude detection and counting of cattle. Ofli \etal \cite{ofli2016combining} release the African Savanna data set as part of their crowd-sourced disaster response project.

\subsection{Labeled Data Sets in Maritime Environments}
Many data sets in maritime environments are captured from satellite-based synthetic aperture radar and therefore fall into the remote sensing category. In this category, the airbus ship data set \cite{airbus-ship} is prominent, featuring 40k images from synthetic aperture radars with instance segmentation labels. Li \etal \cite{li2018hsf} provide a data set of ships with images mainly taken from Google Earth, but also a few UAV-based images. In \cite{xia2018dota}, the authors provide satellite-based images from natural scenes, mainly land-based but also harbors.
The most similar to our work is \cite{lygouras2019unsupervised}. They also consider the problem of human detection in open water. However, their data mostly contains images close to shores and of swimming pools. Furthermore, it is not publicly available.

\subsection{Multi-Modal Data Sets Captured from UAVs}

UAVDT \cite{du2018unmanned} provides coarse meta data for their object detection and tracking data: every frame is labeled with altitude information (low, medium, high), angle of view (front-view, side-view, bird-view) and light conditions (day, night, foggy). Wu \etal \cite{wu2019delving} manually label VisDrone after its release with the same annotation information for the object detection track. Mid-Air \cite{Fonder2019MidAir} is a synthetic multi-modal data set with images in nature containing precise altitude, GPS, time, and velocity data but without annotated objects. Blackbird \cite{antonini2018blackbird} is a real-data indoor data set for agile perception also featuring these meta information. In \cite{majdik2017zurich}, street-view images with the same meta data are captured to benchmark appearance-based localization. Bozcan \etal \cite{bozcan2020air} release a low-altitude ($<30$ m) object detection data set containing images showing a traffic circle and provide meta data such as altitude, GPS, and velocity but exclude the import camera angle information.

Tracking data sets often provide meta data (or attribute information) for the clips. However, in many cases these do not refer to the environmental state in which the image was captured. Instead, they abstractly describe the way in which a clip was captured: UAV123 \cite{mueller2016benchmark} label their clips with information such as aspect ratio change, background clutter, and fast motion, but do not provide frame-by-frame meta data. The same observation can be made for the tracking track of VisDrone \cite{fan2020visdrone}. See Table \ref{table:comparison_datasets} for an overview of annotated aerial data sets.

%------------------------------------------------------------------------
\section{Data Set Generation}

%-------------------------------------------------------------------------
%\subsection{Image Data Collection}
\label{sec:imagedatacollection}

%We gathered the footage on several days to obtain variance in light conditions. Taking into account safety and environmental regulations, we asked over 20 test subjects to be recorded in open water. Only subjects who met strict criteria regarding their ability to swim in open water were recruited. Boats were rented to transport the subjects to the area of interest, where quadcopters were launched at a safe distance from the swimmers. At the same time, the fixed-wing UAV Trinity F90+ was launched from the shore. We used waypoints to ensure a strict flight schedule to maximize data collection efficiency. Care was taken to maintain a strict vertical separation at all times. Subjects were free to wear life jackets, of which we provided several differently colored pieces (see also Figure \ref{fig:objects_examples}).

We gathered the footage on several days to obtain variance in light conditions. Taking into account safety and environmental regulations, we asked over 20 test subjects to be recorded in open water. Boats transported the subjects to the area of interest, where quadcopters were launched at a safe distance from the swimmers. At the same time, the fixed-wing UAV Trinity F90+ was launched from the shore. We used waypoints to ensure a strict flight schedule to maximize data collection efficiency. Care was taken to maintain a strict vertical separation at all times. Subjects were free to wear life jackets, of which we provided several differently colored pieces (see also Figure \ref{fig:objects_examples}).

To diminish the effect of camera biases within the data set, we used multiple cameras, as listed in Table \ref{table:cameras}, mounted to the following drones: DJI Matrice 100, DJI Matrice 210, DJI Mavic 2 Pro, and a Quantum Systems Trinity F90+.
\begin{table}	
	\begin{center}		
		\begin{tabular}{ccc}
			Camera  & Resolution & Video   \\
			\hline			
			Hasselblad L1D-20c & 3,840$\times$2,160 & 30 fps \\
			MicaSense RedEdge-MX & 1,280$\times$ 960 \ \  & \ding{53} \\
			Sony UMC-R10C & 5,456$\times$3,632 & \ding{53} \\
			Zenmuse X5 & 3,840$\times$2,160 & 30 fps \\
			Zenmuse XT2 & 3,840$\times$2,160 & 30 fps
		\end{tabular}
	\end{center}
	\caption{Overview of used cameras.}
	\label{table:cameras}
\end{table}
 With the video cameras, we captured videos at 30 fps. For the object detection task, we extract at most three frames per second of these videos to avoid having redundant occurrences of frames.
 See Section \ref{sec:datasetstatistics} for information on the distribution of images with respect to different cameras.
 
 Lastly, we captured top-down looking multi-spectral imagery at 1 fps. We used a MicaSense RedEdge-MX, which records five wavelengths (475 nm, 560 nm, 668 nm, 717 nm, 842 nm). Therefore, in addition to the RGB channels, the recordings also contain a RedEdge and a Near Infrared channel. The camera was referenced with a white reference before each flight. As the RedEdge-MX captures every band individually, we merge the bands using the development kit provided by MicaSense.

%-------------------------------------------------------------------------
\subsection{Meta Data Collection}

\begin{table}	
	\begin{center}		
		\begin{tabular}{cccc}
			Data  & Unit & Min. value & Max.value  \\
			\hline
			%$300,000$
			Time since start & ms & 0 & $\infty$ \\
			Date and Time & ISO 8601 & -- & -- \\
			Latitude & degrees & $-90$ & $+90$ \\
			Longitude & degrees & $-90$ & $+90$ \\
			Altitude & meters & $0$ & $\infty$ \\
			Gimbal pitch & degrees & $0$ & 90 \\
			UAV roll & degrees & $-90$ & $+90$ \\
			UAV pitch & degrees & $-90$ & $+90$ \\
			UAV yaw & degrees & $-180$ & $+180$ \\
			$x$-axis speed & m/s & $0$ & $\infty$ \\
			$y$-axis speed & m/s & $0$ & $\infty$ \\
			$z$-axis speed& m/s & $0$ & $\infty$
		\end{tabular}
	\end{center}
	\caption{Meta data that comes with every image/frame.}
	\label{table:meta_data}
\end{table}

Accompanied with every frame there is a meta stamp, that is logged at 10 hertz. To align the video data (30 fps) and the time stamps, a nearest neighbor method was performed. The data in Table \ref{table:meta_data} is logged and provided for every image/frame read from the onboard clock, barometer, IMU and GPS sensor, and the gimbal, respectively.
\iffalse
\begin{itemize}
	\itemsep-0.5em
	\item $d$: current date and time of capture
	\item $t$: relative time stamp since beginning of capture
	\item $la$: latitude of the UAV
	\item $lo$: longitude of the UAV
	\item $a$: altitude of the UAV
	\item $\alpha$: camera pitch angle (viewing angle)
	\item $\phi$: UAV roll angle
	\item $\theta$: UAV pitch angle
	\item $\psi$: UAV yaw angle 
	\item $V_x$: speed along the $x$-axis
	\item $V_y$: speed along the $y$-axis
	\item $V_z$: speed along the $z$-axis
\end{itemize}
\fi

 Note that $\alpha=90^\circ$ corresponds to a top-down view, and $\alpha=0^\circ$ to a horizontally facing camera. The date format is given in the extended form of ISO 8601.
Furthermore, note that the UAV roll/pitch/yaw-angles are of minor importance for meta-data-aware vision-based methods as the onboard gimbal filters out movement by the drone such that the camera pitch angle is roughly constant if it is not intentionally changed  \cite{jkedrasiak2013prototype}. Note that the gimbal yaw angle is not included, as we fix it to coincide with the UAV's yaw angle.

We need to emphasize that the meta values lie within the error thresholds introduced by the different sensors, but an extended analysis is beyond the scope of this paper (see \eg \cite{zimmermann2017precise,webinar2020gps,kulhavy2017accuracy} for an overview).

%-------------------------------------------------------------------------
\subsection{Annotation Method}

Using the non-commercial labeling tool DarkLabel \cite{darklabel}, we manually and carefully annotated all provided images and frames with the categories swimmer (person in water without life jacket), floater (person in water with life jacket), life jacket, swimmer$^\dagger$ (person on boat without life jacket), floater$^\dagger$ (person on boat with life jacket), and boats. We note that it is not sufficient to infer the class floater by the location from swimmer and life jacket as this can be highly ambiguous. Subsequently, all annotations were checked by experts in aerial vision. We choose these classes as they are the hardest and most critical to detect in SAR missions. Furthermore, we annotated regions with other objects as ignored regions, such as boats on land. Moreover, the  data  set  also  covers unlabeled objects, which may not be of interest, like driftwood, birds or the coast such that detectors can be robust to distinguish from those objects. Our guidelines for the annotation are described in the appendix. See Figure \ref{fig:objects_examples} for examples of objects.

%In particular, swimmers were annotated such that the complete body including arms and legs (if visible) are within the bounding box. The bounding box format is $(x,y,w,h)$, where $x$ and $y$ correspond to the upper left corner and $w$ and $h$ to the width and height, respectively. 

\begin{figure*}
	\centering

	\setlength{\tabcolsep}{2pt}
			\begin{tabular}{ccccc}
	            
					% left, bottom, right, top

				%{\includegraphics[width=31mm,height=30mm,trim=0 0 0 0,clip]{floater_example1_l.png}}
				
				\begin{overpic}[width=31mm,height=30mm,tics=10]{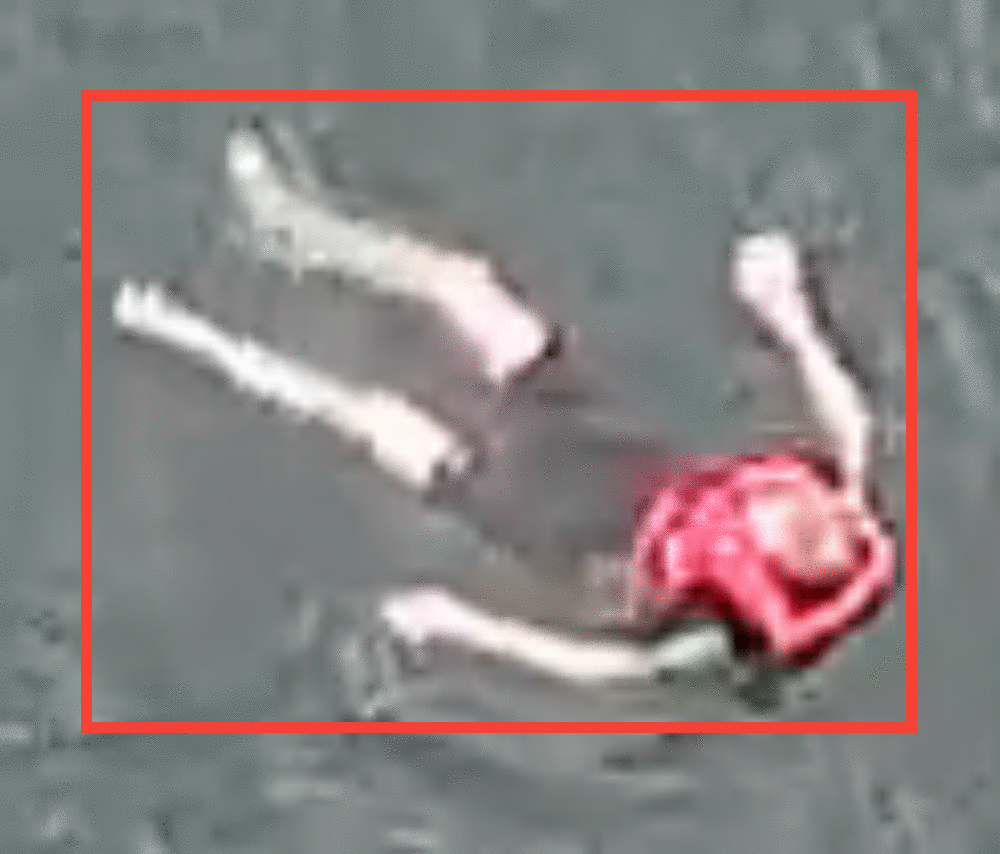} \put (0,4) {\large \colorbox{blue!30}{Floater}}\end{overpic}				
			
				&

				%{\includegraphics[width=31mm,height=30mm,trim=0 0 0 0,clip]{floater_example2_l.png}}
				
				\begin{overpic}[width=31mm,height=30mm,tics=10]{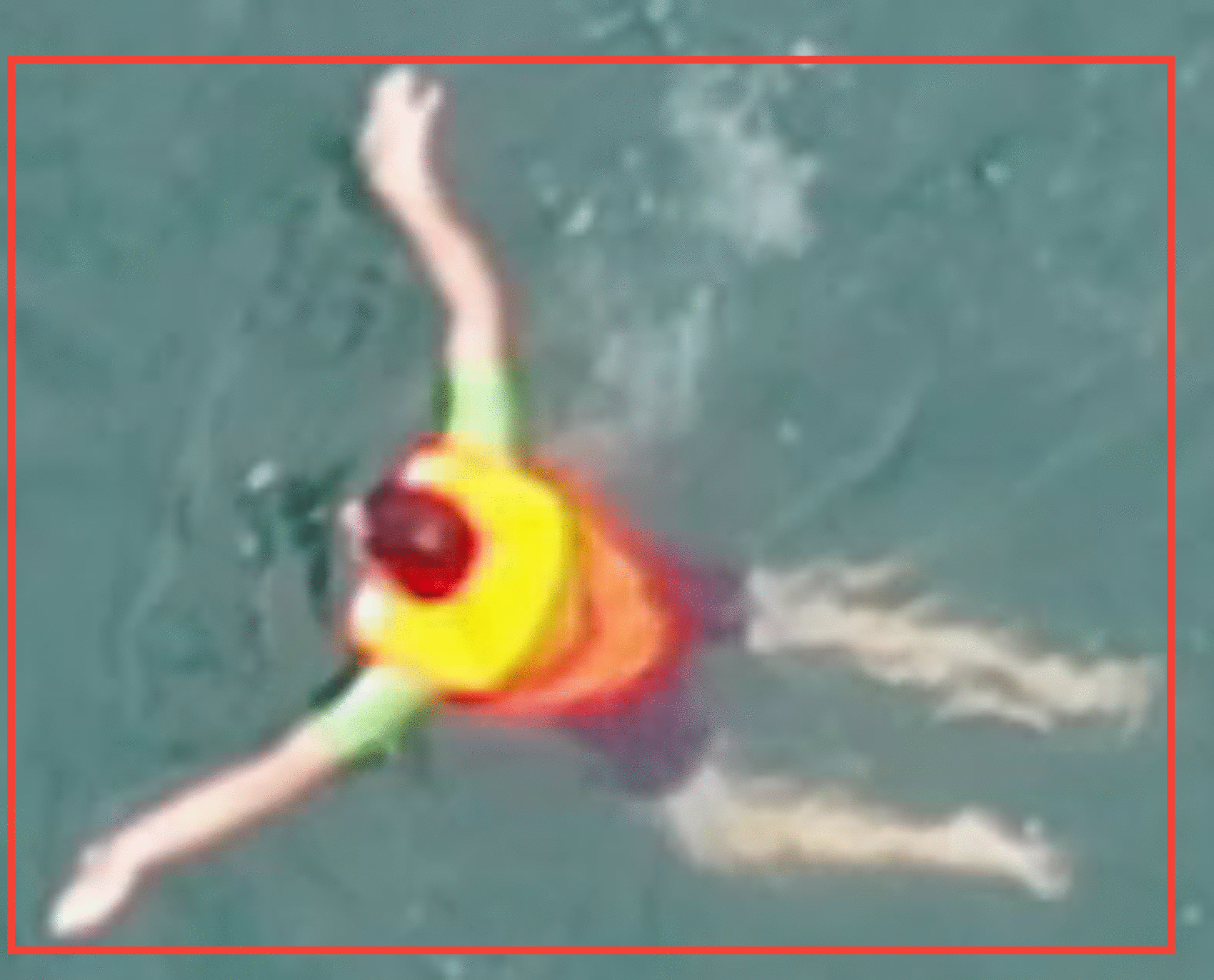} \put (0,4) {\large \colorbox{blue!30}{Floater}}\end{overpic}

				\hphantom{abc}
				&
				%{\includegraphics[width=31mm,height=30mm,trim=0 0 0 0,clip]{floater_example3_l.png}}&

				\begin{overpic}[width=31mm,height=30mm,tics=10]{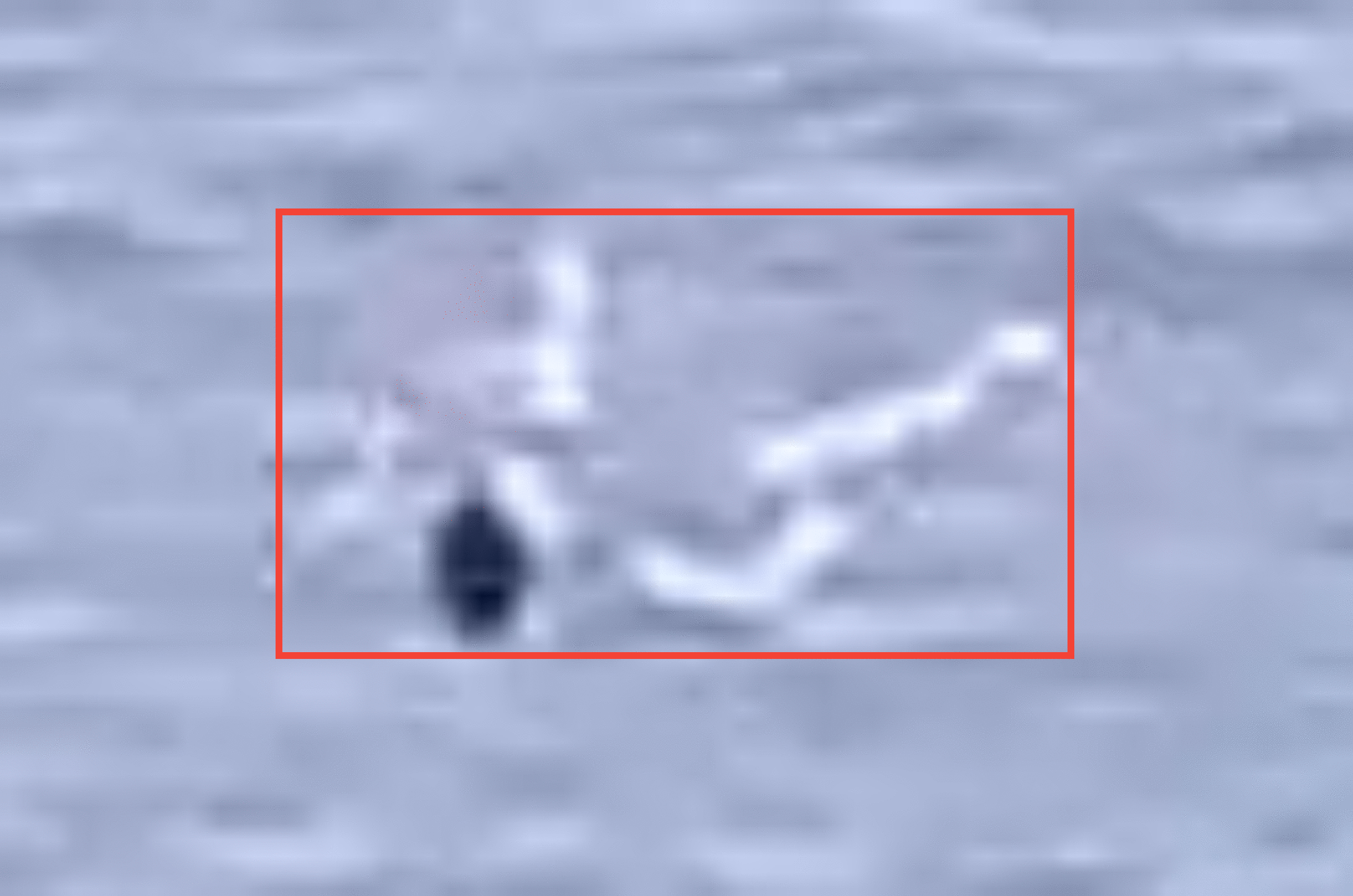} \put (0,4) {\large \colorbox{blue!30}{Swimmer}}\end{overpic}	
				
				%{\includegraphics[width=31mm,height=30mm,trim=0 0 0 0,clip]{floater_example4_l.png}}
				
				\begin{overpic}[width=31mm,height=30mm,tics=10]{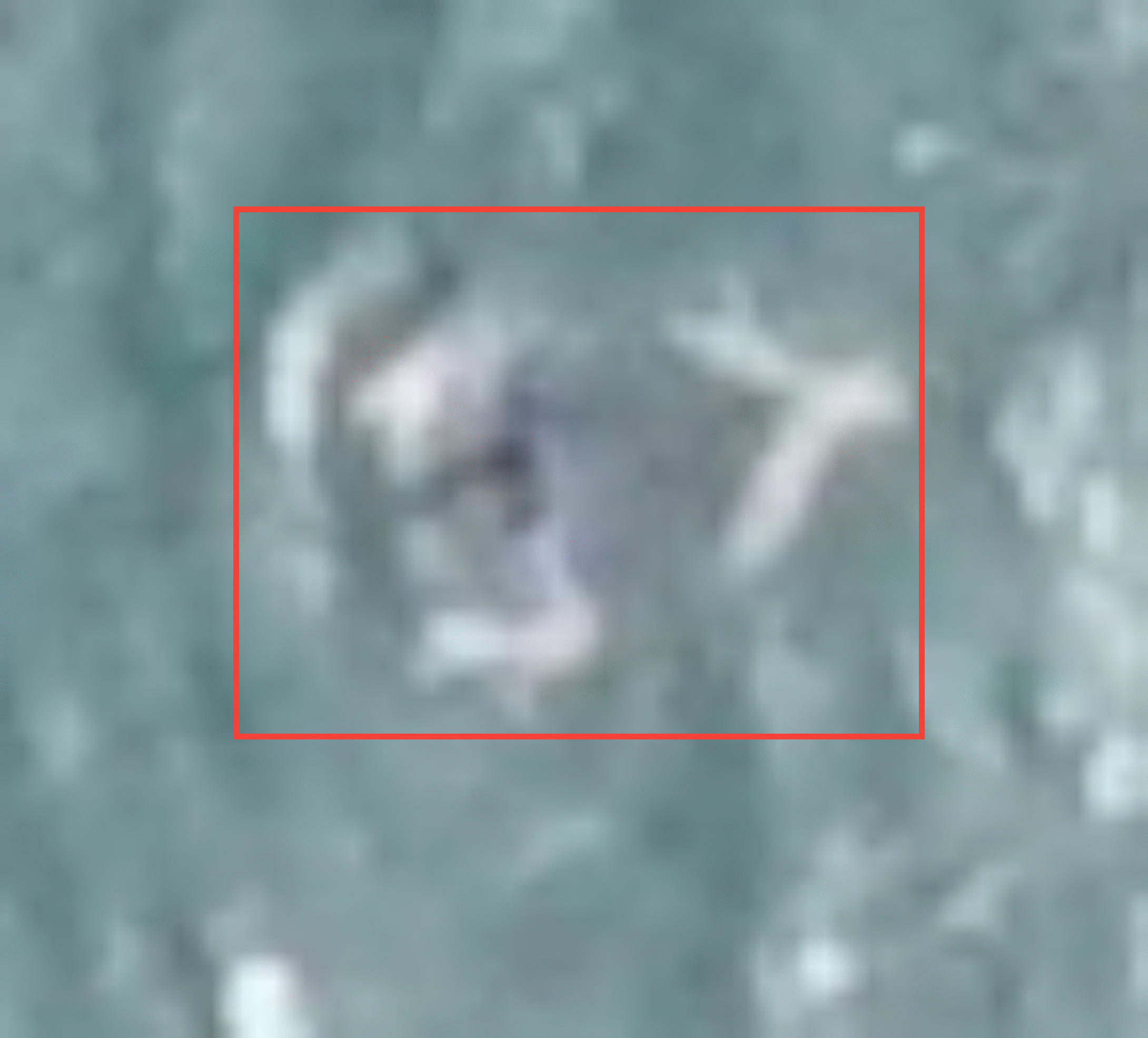} \put (0,4) {\large \colorbox{blue!30}{Swimmer}}\end{overpic}
				
				\hphantom{abc} &
				
				%{\includegraphics[width=31mm,height=30mm,trim=0 0 0 0,clip]{human_on_boat_with_jacket_l.png}}
				
				\begin{overpic}[width=31mm,height=30mm,tics=10]{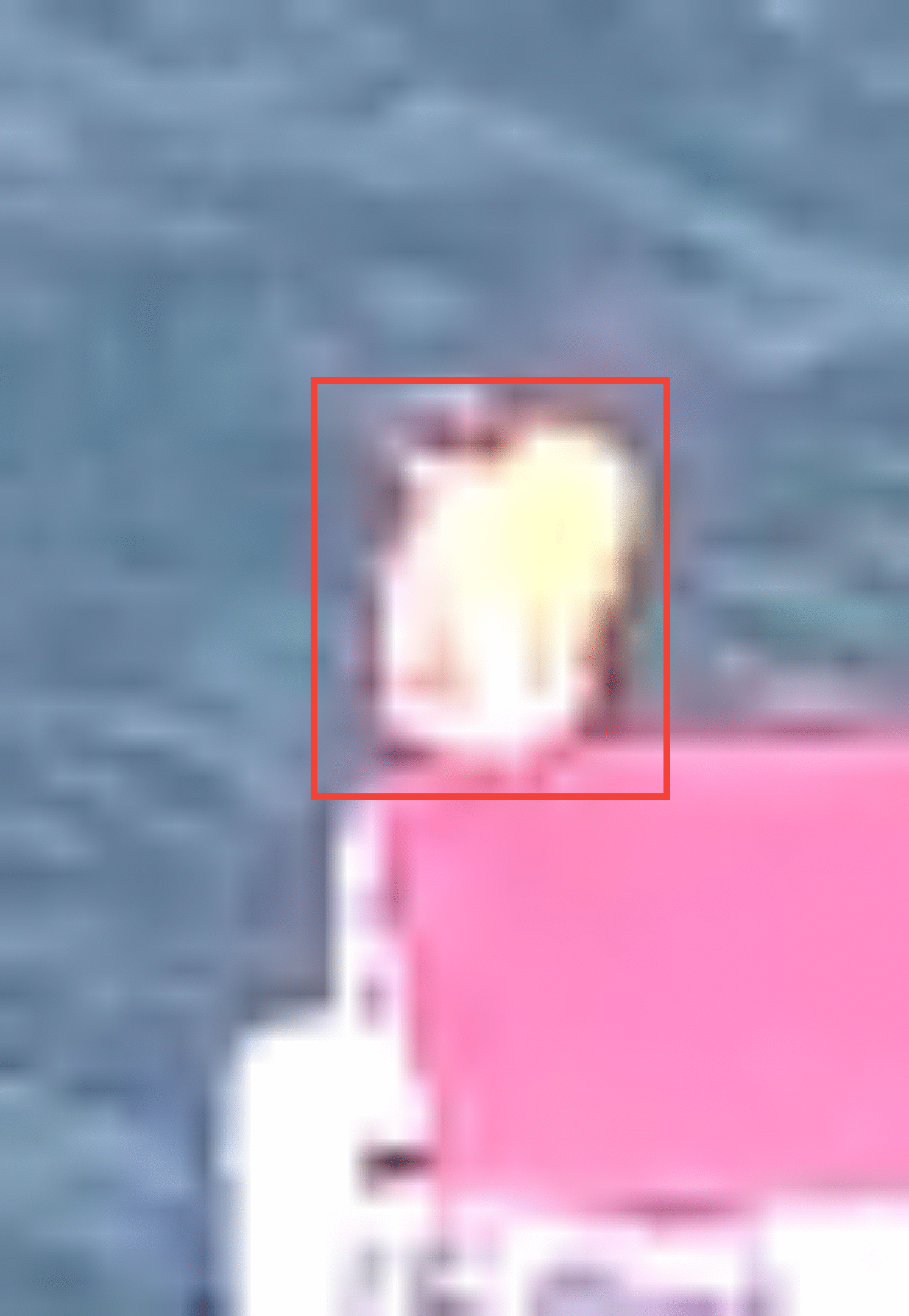} \put (0,4) {\large \colorbox{blue!30}{Floater$^\dagger$}}\end{overpic}
				
				\\

			%{\includegraphics[width=31mm,height=30mm,trim=0 300 0 300,clip]{floater_example5_l.png}}
			
			\begin{overpic}[width=31mm,height=30mm,tics=10]{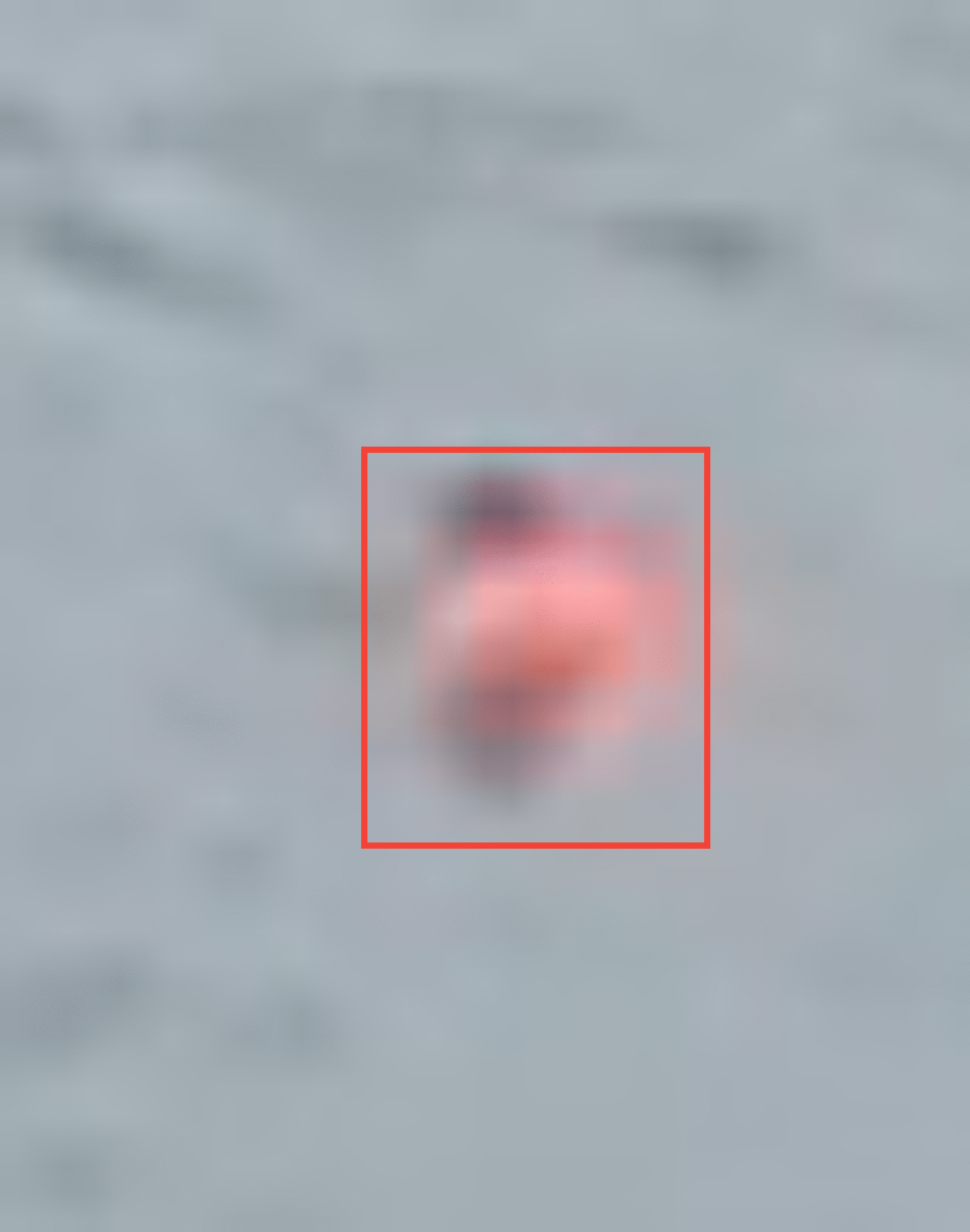} \put (0,4) {\large \colorbox{blue!30}{Floater}}\end{overpic}			
			
			&
			
			%{\includegraphics[width=31mm,height=30mm,trim=0 0 0 0,clip]{swimmer_example3.png}}&
			%{\includegraphics[width=31mm,height=30mm,trim=0 0 0 0,clip]{life_jacket_l.png}}
			
			\begin{overpic}[width=31mm,height=30mm,tics=10]{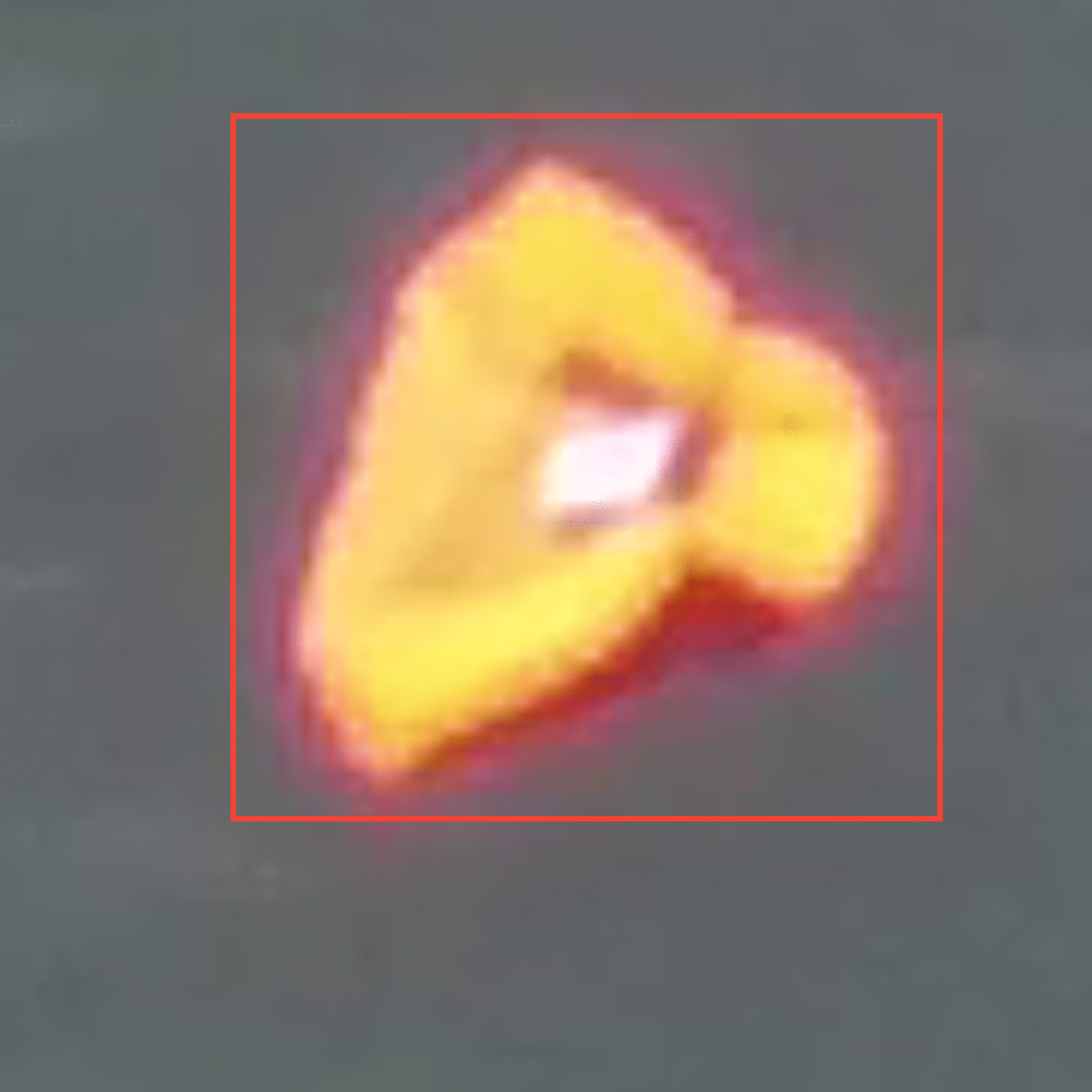} \put (0,6) {\large \colorbox{blue!30}{Life jacket}}\end{overpic}

			\hphantom{abc}
			
			%{\includegraphics[width=31mm,height=30mm,trim=0 0 0 0,clip]{swimmer_example4_l.png}}
			
			&
			
			\begin{overpic}[width=31mm,height=30mm,tics=10]{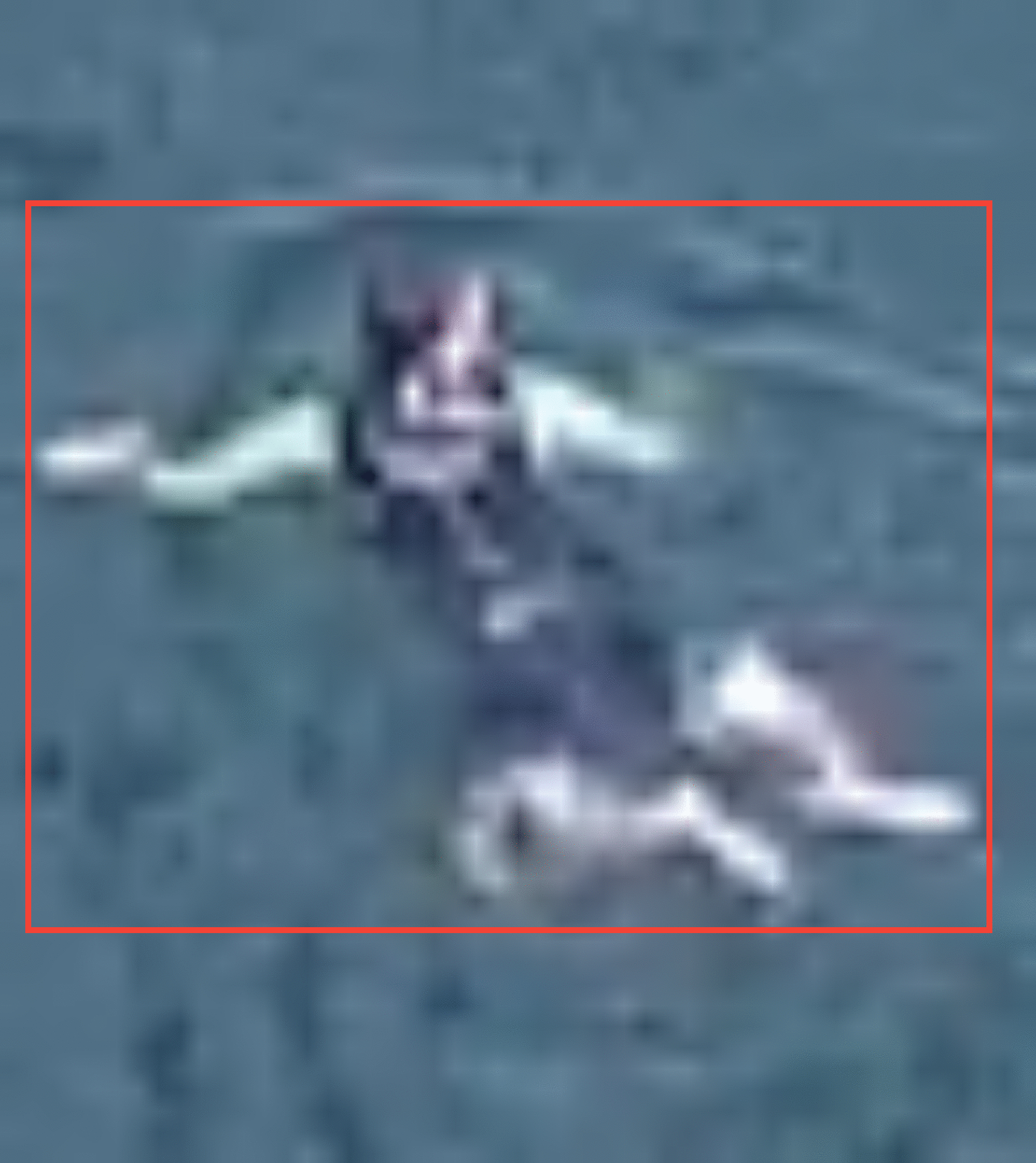} \put (0,3) {\large \colorbox{blue!30}{Swimmer}}\end{overpic}

			%{\includegraphics[width=31mm,height=30mm,trim=0 0 0 0,clip]{swimmer_example2_l.png}}
			
			\begin{overpic}[width=31mm,height=30mm,tics=10]{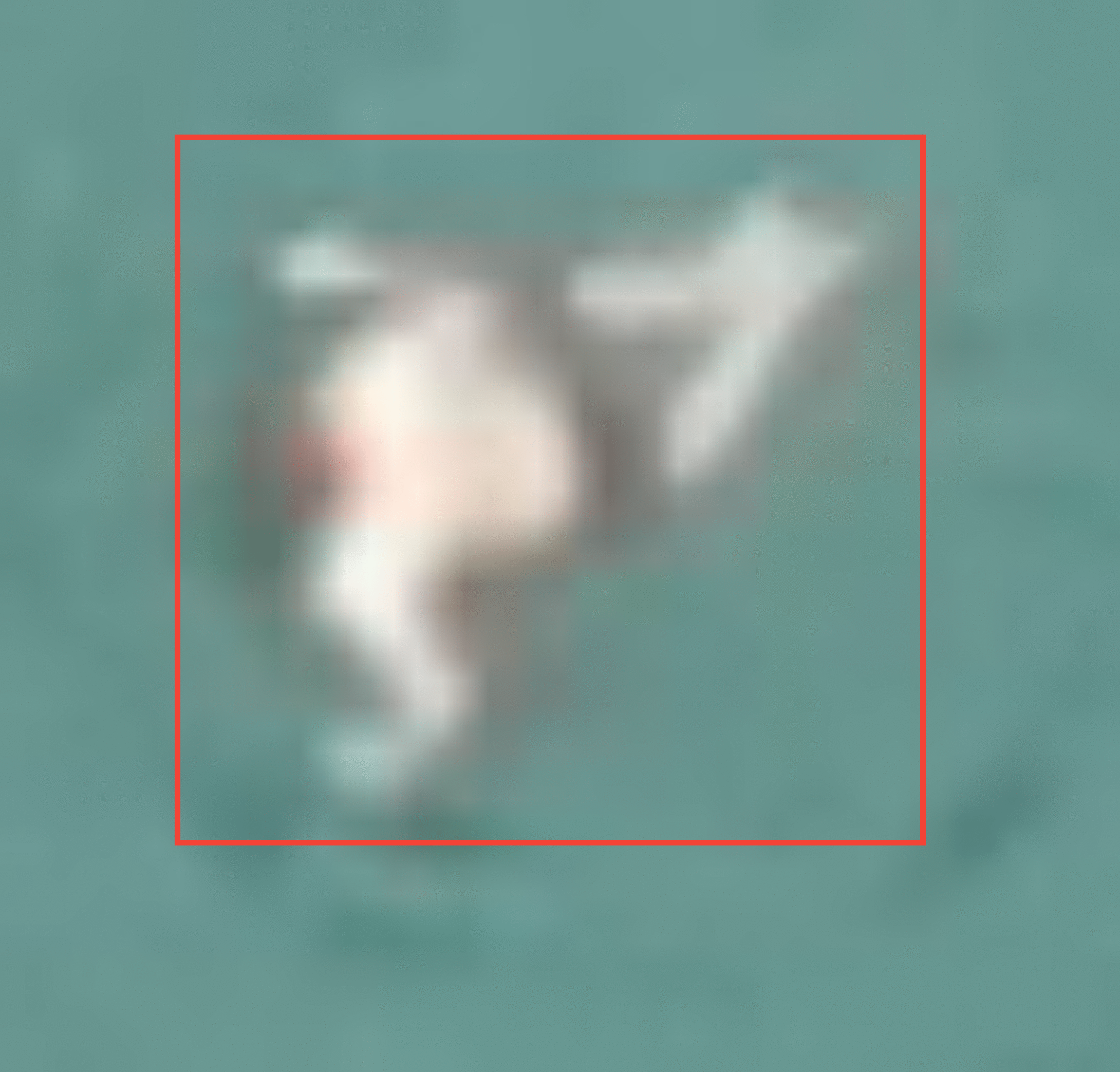} \put (0,4) {\large \colorbox{blue!30}{Swimmer}}\end{overpic}
				
			\hphantom{abc} &
			%{\includegraphics[width=31mm,height=30mm,trim=0 0 0 0,clip]{human_on_boat_better.png}}
			\begin{overpic}[width=31mm,height=30mm,tics=10]{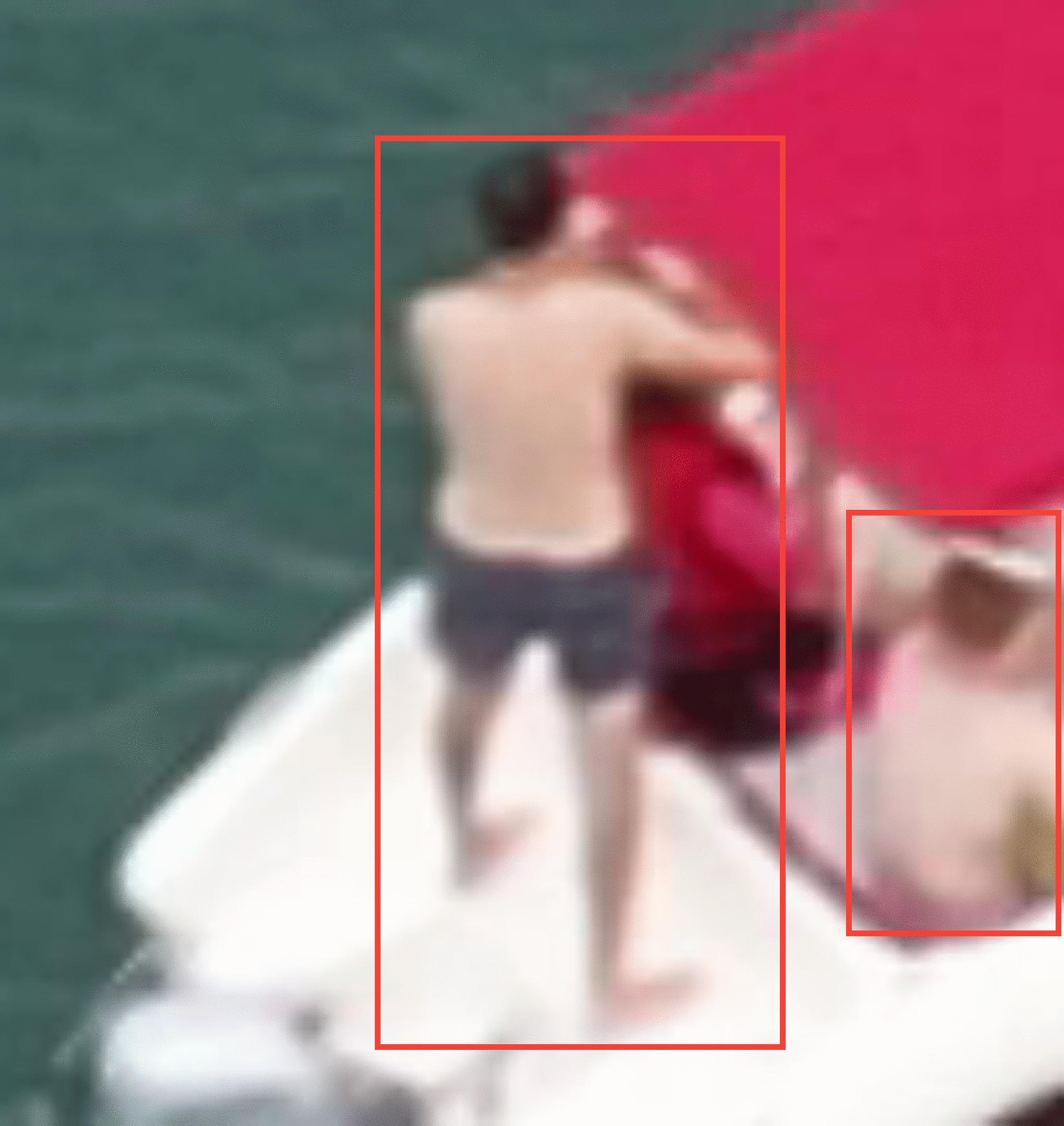} \put (0,3) {\large \colorbox{blue!30}{Swimmers$^\dagger$}}\end{overpic}
			\end{tabular}

	\caption{Examples of objects. Note that these examples are crops from high-resolution images. However, as the objects are small and the images taken from high altitudes, they appear blurry.\protect\\
}
	\label{fig:objects_examples}
	\vspace{-0.6cm}

\end{figure*}

%-------------------------------------------------------------------------
\subsection{Data Set Split}
\label{sec:datasetsplit}

\subsubsection*{Object Detection}
To ensure that the training, validation, and testing set have similar statistics, we roughly balance them such that the respective subsets have similar distributions with respect to altitude and angle of view, two of the most important factors of appearance changes. Of the individual images, we randomly select $\nicefrac{4}{7}$ and add it to the training set, add $\nicefrac{1}{7}$ to the validation set and another $\nicefrac{2}{7}$ to the testing set. In addition to the individual images, we randomly cut every video into three parts of length $\nicefrac{4}{7}$, $\nicefrac{1}{7}$, and $\nicefrac{2}{7}$ of the original length and add every 10-th frame of the respective parts to the training, validation, and testing set. This is done to avoid having subsequent frames in the training and testing set such that a realistic evaluation is possible. We release the training and validation set with all annotations and the testing set's images, but withhold its annotations. Evaluation will be available via an evaluation server, where the predictions on the test set can be uploaded.
\vspace*{-3mm}

%TODO:
%videos for tracking with different meta data distribution
%do NOT write frome same video
%think this is done

\subsubsection*{Object Tracking}

Similarly, we take $\nicefrac{4}{7}$ of our recorded clips as the training clips, $\nicefrac{1}{7}$ as the validation clips and $\nicefrac{2}{7}$ as the testing clips. As for the object detection task, we withhold the annotations for the testing set and provide an evaluation server.

%The tracking clips are based on the already described video splits, i.e. the training set consists of every recorded sequence's first $\nicefrac{4}{7}$ and so on. Hence the sets are disjoint. Both single-object tracking and multi-object tracking use the same clips. Like for the object detection task, we withhold the annotations for the testing set and provide an evaluation server for both, single-object tracking and multi-object tracking.

%-------------------------------------------------------------------------
\section{Data Set Tasks}
\label{sec:datasetstatistics}

There are many works on UAV-based maritime SAR missions, focusing on unified frameworks describing the process of how to search and rescue people \cite{mishra2020drone,gallego2019detection,lvsouras2020new,lygouras2019unsupervised,queralta2020autosos,roberts2016unmanned,ghazali2016determining}. These works answer questions corresponding to path planning, autonomous navigation and efficient signal transmission. Most of them rely on RGB sensors and detection and tracking algorithms to actually find people of interest. This commonality motivates us to extract the specific tasks of object detection and tracking, which pose some of the most challenging issues in this application scenario.

Maritime environments from a UAV's perspective are difficult for a variety of reasons: Reflective regions and shadows resulting from different cardinal points (such as in Fig. \ref{fig:front_image}) that could lead to false positives or negatives; people may be hardly visible or occluded by waves or sea foam (see Supplementary material); typically large areas are overseen such that objects are particularly small \cite{mishra2020drone}. We note that these factors are on top of general UAV-related detection difficulties.

Now, we proceed to describe the specific tasks.

%TODO passt so?
%glaub schon

%What is common among these approaches is to use RGB sensors and detection or tracking algorithms to actually find
%As motivated in the introduction, we focus on the detection, localization, and tracking of people in open water. This is mainly due to works providing many unified frameworks that. We would like to emphasize that 

%challenges in maritime
%The   most   common challenges  for  human  in  marine  environment  detection  are the  following:  the  background  movements  at  the  reflective regions  and  shadows  that  might  be  easily  miss-identified  as foreground  objects’  movements  for  human  detection.  In addition, poor visibility of swimmers due to reflections (from sunlight  and  night-time  lighting)  and  the  issue  of  occlusion make  accurate  segmentation  a  very  challenging  task.  Apart from  the  above  unique  issues  in  aquatic  environment,  there are  common  challenges  faced  in  outdoor  surveillance  (i.e., continual illumination changes due to ambient lighting, auto-gain effects of the cameras, etc.). Fast background updating is important  to  adapt  to  such  illumination  changes.
%sonnenreflektion,sonnenstand
%untergrundbewegung durch wellen
%wellen verdecken objekte

%-------------------------------------------------------------------------
\subsection{Object Detection}
\label{sec:od_statistics}

\begin{figure}
	
	\centering
		% left, bottom, right, top
	\includegraphics[scale=0.5,trim=0 0 0 8,clip]{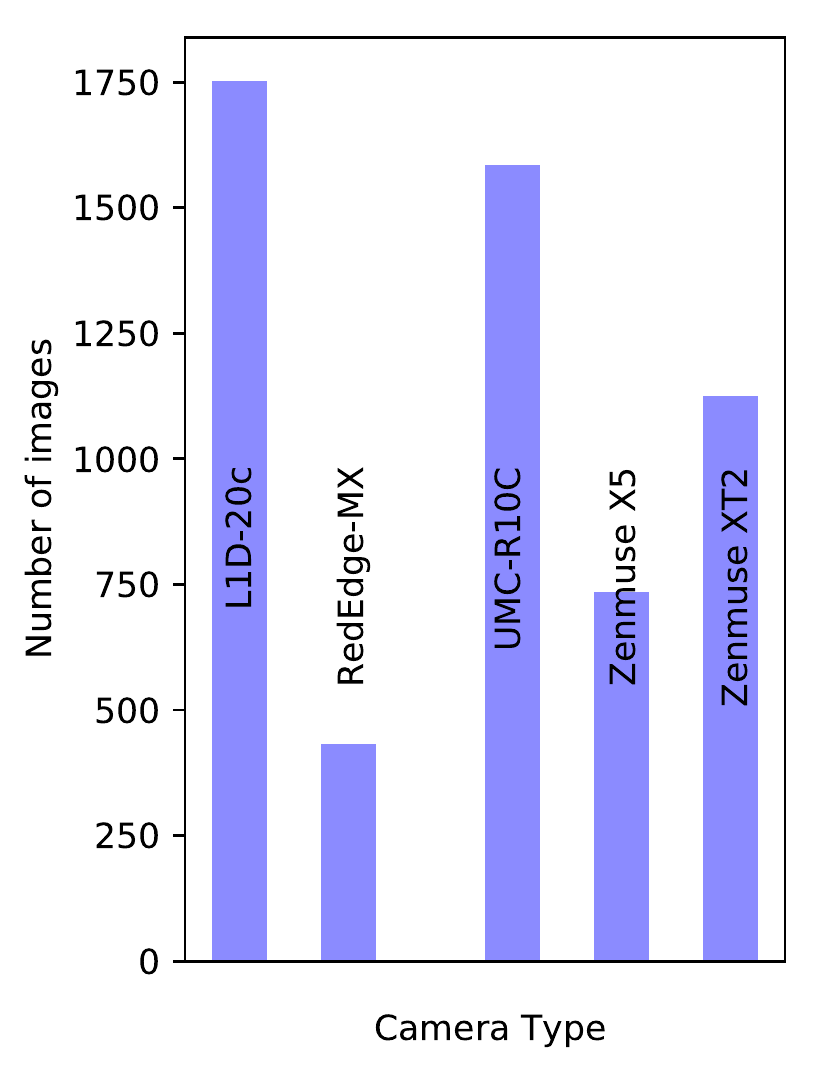}	
	\includegraphics[scale=0.5]{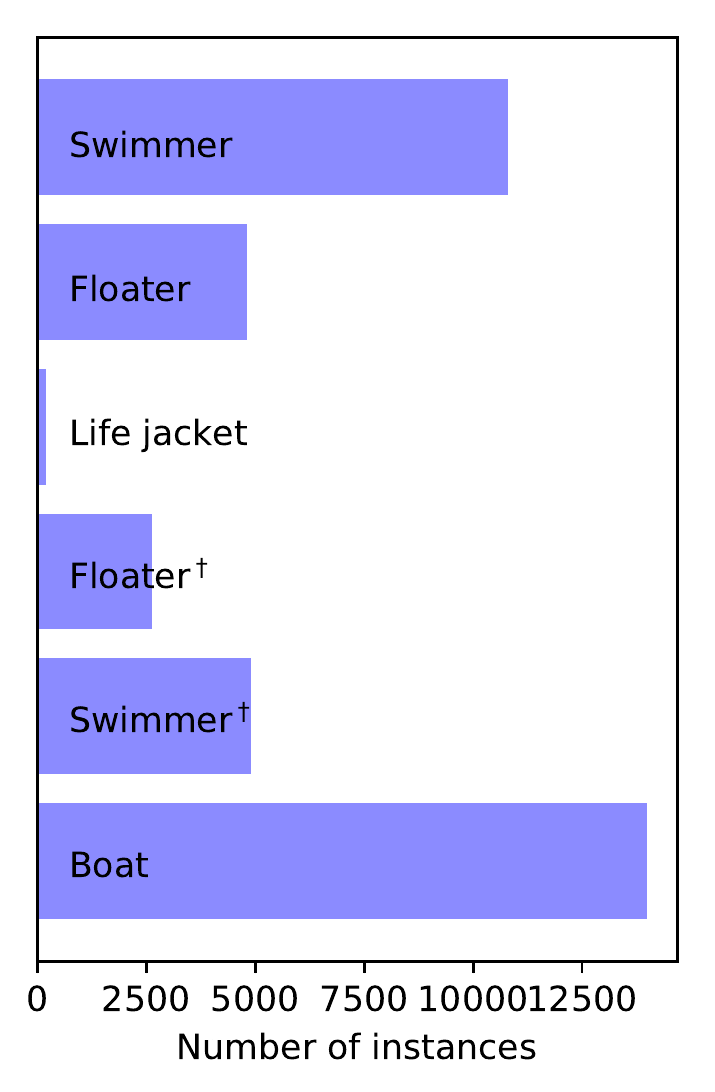}

	\caption{Distribution of training images over camera types (left) and distribution of objects over classes (right).}
	\label{fig:camera_and_class_distribution}
\end{figure}

There are 5,630 images (training: 2,975; validation: 859; testing: 1,796). See Figure \ref{fig:camera_and_class_distribution} for the distribution of images/frames with respect to cameras and the class distribution. We recorded most of the images with the L1D-20c and UMC-R10C, having the highest resolution. Having the lowest resolution, we recorded only 432 images with the RedEdge-MX. Note, for the Object Detection Task only the RGB-channels of the multi-spectral images are used to support a uniform data structure.

 Furthermore, the class distribution is slightly skewed towards the class 'boat', since safety precautions require boats to be nearby. We emphasize that this bias can easily be diminished by blackening the respective regions, as is common for areas which are not of interest or undesired (such as boats here; see \eg \cite{du2018unmanned}). Right after that, swimmers with life jacket are the most common objects. We argue that this scenario is very often encountered in SAR missions. This type of class often is easier to detect than just swimmer as life jackets mostly are of contrasting color, such as red or orange (see Fig. \ref{fig:objects_examples} and Table \ref{table:od_results}). However, as it is also a likely scenario to search for swimmers without life jacket, we included a considerable amount. There are also several different manifestations/visual appearances of that class which is why we recorded and annotated swimmers with and without adequate swimwear (such as wet suit). To be able to discriminate between humans in water and humans on boats, we also annotated humans on boats (with and without life jackets). Lastly, we annotated a small amount of life jackets only. However, we note that the discrimination between life jackets and humans in life jackets can become visually ambiguous, especially in higher altitudes. See also Fig. \ref{fig:objects_examples}.

\begin{figure}
	\vspace*{-3mm}
	\centering
	\includegraphics[scale=0.5]{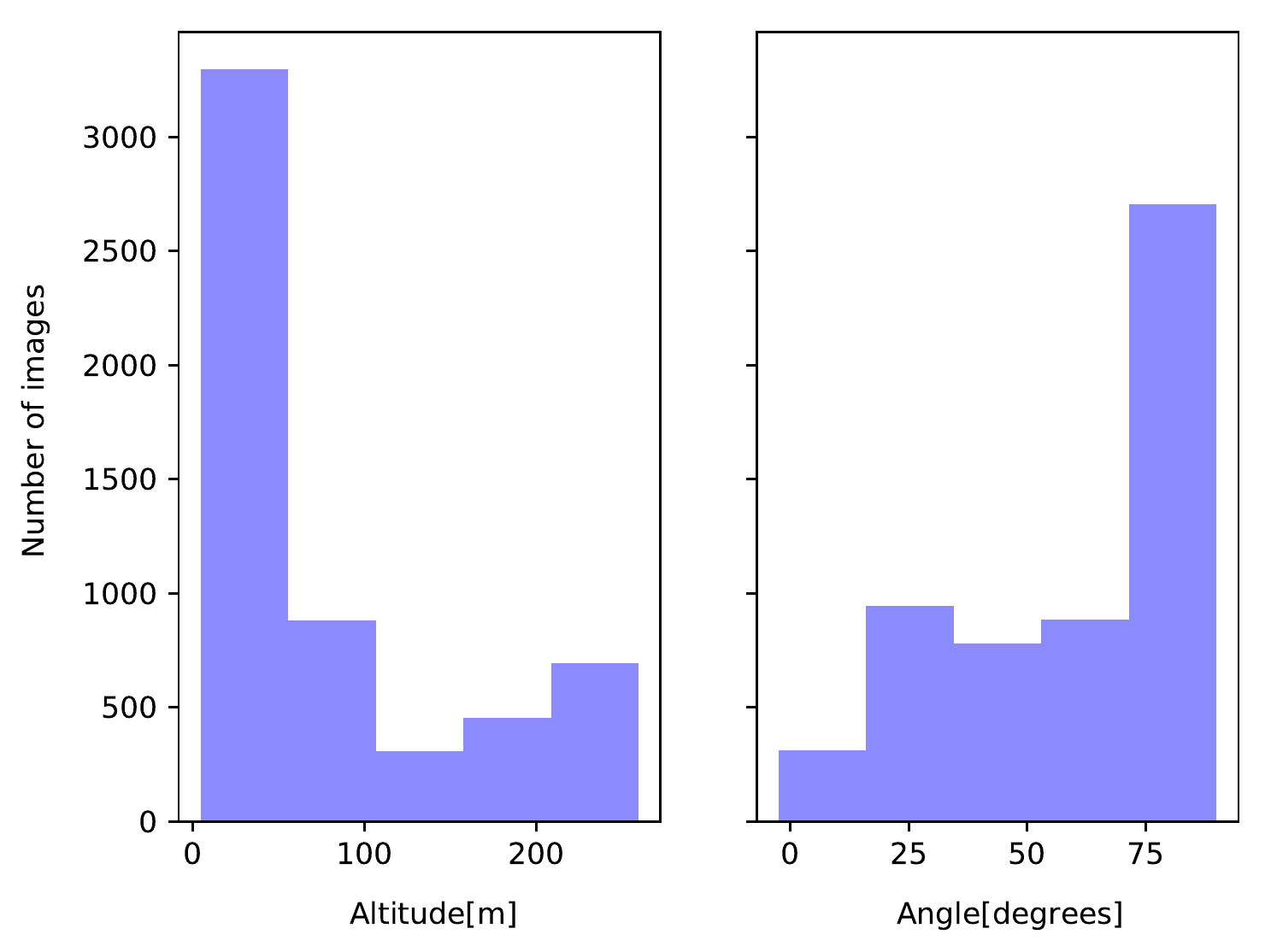}

	\caption{Distribution of images over altitudes (left) and angles (right), respectively.}
	\label{fig:meta_distribution}
	\vspace*{-5mm}
\end{figure}

Figure \ref{fig:meta_distribution} shows the distribution of images with respect to the altitude and viewing angle they were captured at. Roughly 50\% of the images were recorded below 50 m because lower altitudes allow for the whole range of available viewing angles ($0-90^\circ$). That is, to cover all viewing angles, more images at these altitudes had to be taken. On the other hand, there are many images facing downwards ($90^\circ$), because images taken at greater altitudes tend to face downwards since acute angles yield image areas with tiny pixel density, which is unsuitable for object detection. Nevertheless, every altitude and angle interval is sufficiently represented.

%-------------------------------------------------------------------------

\subsection{Single-Object Tracking}

We provide 208 short clips ($>$4 seconds) with a total of 393,295 frames (counting the duplicates), including all available objects labeled. We randomly split the sequences into 58 training, 70 validation and 80 testing sequences. We do not support long-term tracking. The altitude and angle distributions are similar to these in the object detection section since the origin of the images of the object detection task is the same.

\subsection{Multi-Object Tracking}

We provide 22 clips with a total of 54,105 frames and 403,192 annotated instances, the average consists of 2,460 frames. We differentiate between two use-cases. In the first task, only the persons in water (floaters and swimmers) are tracked, it is called MOT-Swimmer. In the second task, all objects in water are tracked (also the boats, but not people on boats), called MOT-All-Objects-In-Water. In both tasks, all objects are grouped into one class. The data set split is performed as described in section \ref{sec:datasetsplit}.
%Like for the Single Object Tracking, we randomly split the sequences into training, validation and testing sequences.

\subsection{Multi-Spectral Footage}
\label{sec:multi_spectral}

Along with the data for the three tasks, we provide multi-spectral images. We supply annotations for all channels of these recordings, but only the RGB-channels are currently part of the Object Detection Task. 
There are 432 images with 1,901 instances. See Figure \ref{fig:front_image} for an example of the individual bands.

%------------------------------------------------------------------------
\section{Evaluations}
\label{sec:evaluations}

We evaluate current state-of-the-art object detectors and object trackers on SeaDronesSee. All experiments can be reproduced by using our provided code available on the evaluation server. Furthermore, we refer the reader to the Supplementary Material for the exact form and uploading requirements.

%------------------------------------------------------------------------
\subsection{Object Detection}
\label{sec:evaluation_od}

%\multicolumn{1}{c!{\vline height 1.1\ht\strutbox}}{Model}  & AP & AP$_{50}$ & AP$_{75}$ & AR$_{1}$ & \multicolumn{1}{c!{\vline height 1.1\ht\strutbox}}{AR$_{10}$}  & \begin{turn}{90}swimmer\end{turn} & \begin{turn}{90}floater\end{turn} & \begin{turn}{90} swimmer^\dagger\end{turn} & \begin{turn}{90}floater^\dagger\end{turn} & \begin{turn}{90}boat\end{turn} & \multicolumn{1}{c!{\vline height 1.1\ht\strutbox}}{\begin{turn}{90}life jacket\end{turn}} & FPS \\

%\newcommand{degree}{90}
\begin{table*}[h!]	
	\begin{center}		
		\begin{tabular}{c|ccccc|cccccc|c}
			\multicolumn{1}{c!{\vline height 1.1\ht\strutbox}}{Model}  & AP & AP$_{50}$ & AP$_{75}$ & AR$_{1}$ & {AR$_{10}$}  & S & F & S$^\dagger$ & F$^\dagger$ & B & LJ & FPS \\
			\hline
			F. ResNeXt-101-FPN \cite{xie2017aggregated}  & 30.4 & 54.7 & 29.7 & 18.6 & 42.6 & 78.1 & 82.4 & 25.9 & 44.3 & 96.7 & 0.6 & 2\\
			F. ResNet-50-FPN \cite{girshick2015fast} & 14.2 & 30.1 & 7.2 & 6.4 & 17.7 & 24.6 & 54.1 & 4.9 & 7.5 & 89.2 & 0.3 & 14 \\ \hline
			CenterNet-Hourglass104 \cite{zhou2019objects} & 25.6 & 50.3 & 22.2 & 17.7 & 40.1 & 65.1 & 73.6 & 19.1 & 48.1 & 95.8 & 0.3 & 6 \\
			CenterNet-ResNet101 \cite{zhou2019objects} & 15.1 & 36.4 & 10.8 & 9.6 & 21.4 & 16.8 & 39.8 & 0.8 & 1.7 & 74.3 & 0 &  22 \\
			%CenterNet-ResNet50 \cite{zhou2019objects} & 15.9 & 39.9 & 9.6 & 11.3 & 23.5 & 12.2 & 19.3 & 1.2 & 7.8 & 56.8 & 0 & 33 \\
			CenterNet-ResNet18 \cite{zhou2019objects} & 9.9 & 21.8 & 9.0 & 7.2 & 19.7 & 20.9 & 21.9 & 2.6 & 3.3 & 81.9 & 0.4 & 78 \\ 
			\hline
			EfficientDet--$D0$ \cite{tan2020efficientdet} & 20.8 & 37.1 & 20.6 & 11.5 & 29.1 & 65.3 & 55.1 & 3.1 & 3.3 & 95.5 & 0.1 & 26
			% EfficientDet--$D2$ \cite{tan2020efficientdet} & 20.1 & 33.5 & \textsc{tbd} & 10.5 & 25.3 & 60.9 & 43.8 & 1.0 & 0.6 & 94.6 & 0.0 & \textsc{tbd}

		\end{tabular}
	\end{center}
	\caption{Average precision results for several baseline models. The right part contains AP$_{50}$--values for each class individually. All reported FPS numbers are obtained on a single Nvidia RTX 2080 Ti. The abbreviation 'F.' stands for Faster R-CNN. For visualization purposes, the classes are abbreviated as swimmer($^\dagger$) $\rightarrow$ S($^\dagger$), floater($^\dagger$) $\rightarrow$ F($^\dagger$), boat $\rightarrow$ B, life jacket $\rightarrow$ LJ. }
	\label{table:od_results}
	\vspace{-0.4cm}

\end{table*}
%%%%%%%
%test: can be deleted

%\iffalse

%\fi

The used detectors can be split into two groups. The first group consists of two-stage detectors, which are mainly built on Faster R-CNN \cite{girshick2015fast} and its improvements. Built for optimal accuracy, these models often lack the inference speed needed for real-time employment, especially on embedded hardware, which can be a vital use-case in UAV-based SAR missions. For that reason, we also evaluate on one-stage detectors. In particular, we perform experiments with the best performing single-model (no ensemble) from the workshop report \cite{zhu2018visdrone}: a Faster R-CNN with a ResNeXt-101 64-4d \cite{xie2017aggregated} backbone with P6 removed. For large one-stage detectors, we take the recent CenterNet \cite{zhou2019objects}. To further test an object detector in real-time scenarios, we choose the current best model family on the COCO test-dev according to \cite{paperswithcode}, i.e. EfficientDet \cite{tan2020efficientdet}, and take the smallest model, $D0$, which can run in real-time on embedded hardware, such as the Nvidia Xavier \cite{kiefer2021leveraging}. We refer the reader to the appendix for the exact parameter configurations and training configurations of the individual models.

Similar to the VisDrone benchmark \cite{zhu2018vision}, we evaluate detectors according to the COCO json-format \cite{lin2014microsoft}, i.e. average precision at certain intersection-over-union-thresholds. More specifically, we use AP$=$AP$^{\text{IoU=0.5:0.05:0.95}}$, AP$_{50}=$AP$^{\text{IoU=0.5}}$ and AP$_{75}=$AP$^{\text{IoU=0.75}}$. Furthermore, we evaluate the maximum recalls for at most 1 and 10 given detections, respectively, denoted AR$_1=$AR$^{\text{max=1}}$, and AR$_{10}=$AR$^{\text{max=10}}$. All these metrics are averaged over all categories (except for "ignored region"). We furthermore provide the class-wise average precisions. 
Moreover, similar to \cite{kiefer2021leveraging}, we report AP$_{50}$-results on different equidistant levels of altitudes 'low' = 5-56 m (L), 'low-medium' = 55-106 m (LM), 'medium' = 106-157 m (M), 'medium-high' = 157-208 m (MH), and 'high' = 208-259 m (H). To measure the universal cross-domain performance, we report the average over these domains, denoted AP$_{50}^\text{avg}$. Similarly, we report AP$_{50}$-results for different angles of view: 'acute' = 7-23$^\circ$ (A), 'acute-medium' = 23-40$^\circ$ (AM), 'medium' = 40-56$^\circ$ (M), 'medium-right' = 56-73$^\circ$ (MR), and 'right' = 73-90$^\circ$ (R). Ultimately, it is the goal to have robust detectors across all domains uniformly, which is better measured by the latter metrics.
%We refer the reader to \cite{lin2014microsoft} for more details.

Table \ref{table:od_results} shows the results for all object detection models. As expected, the large Faster R-CNN with ResNeXt-101 64-4d backbone performs best, closely followed by CenterNet-Hourglass104. Medium-sized networks, such as the ResNet-50-FPN, and fast networks, such as CenterNet-ResNet18 and EfficientDet-$D0$, expectedly perform worse. However, the latter can run in real-time on an Nvidia Xavier \cite{kiefer2021leveraging}. Swimmers are detected significantly worse than floaters by most detectors. Notably, life jackets are very hard to detect since from a far distance these are easily confused with swimmers$^\dagger$ (see Fig. \ref{fig:objects_examples}). Since there is a heavy class imbalance with many fewer life jackets, detectors are biased towards floaters.

Table \ref{table:domain_aps_altitude} and \ref{table:domain_aps_angle} show the performances for different altitudes and angles, respectively. These evaluations help assess the strength and weaknesses of individual models. For example, although ResNeXt-101-FPN performs overall better than Hourglass104 in AP$_{50}$ (54.7 vs. 50.3), the latter is better in the domain of medium angles (45.2 vs. 49.7). Furthermore, the great performance discrepancy between CenterNet-ResNet101 and CenterNet-ResNet18 in AP$_{50}$ (36.4 vs. 21.8) vanishes when averaged over angle domains  (23.8 vs. 23.1 AP$_{50}^{\text{avg}}$) possibly indicating ResNet101's bias towards specific angle domains.  

%We note, however, that even the ResNeXt-101 64-4d fails occasionally when shown certain wave patterns as shown in Figure \ref{fig:false_positives}. 

\begin{table}

		\resizebox{240pt}{!}{%
			\begin{tabular}{c|ccccc|c}
								Model &  L & LM & M & MH & H & AP$_{50}^{\text{avg}}$ \\ \hline

				ResNeXt-101-FPN & 56.8 & 54.6 & 49.2 & 65 & 78.3 & 60.8\\ 
				ResNet-50-FPN & 32.8 & 29.8 & 23.5 & 40.5 & 48.9 & 35.1 \\ \hline
				Hourglass104 & 50.6 & 52.0 & 47.5 & 64.9 & 73.2 & 57.6\\
				ResNet101 & 20.2 & 30.4 & 24.1 & 35.1 & 38.0 & 29.6\\
				ResNet18 & 23.8 & 20.3 & 19.2 & 29.3 & 31.9 & 24.9\\ \hline
				$D0$ & 39.6 & 38.0 & 30.4 & 42.5 & 54.5 & 41.0
				
		\end{tabular}}

	\caption{Results on different altitude-domains. E.g. ResNeXt's AP$_{50}$ performance in low-medium (LM) altitudes is 54.6 AP$_{50}$.}
	\label{table:domain_aps_altitude}

%\fi
\vspace*{3mm}
%\iffalse

		\resizebox{240pt}{!}{%
			\begin{tabular}{c|ccccc|c}
								Model &  A & AM & M & MR & R & AP$_{50}^{\text{avg}}$ \\ \hline

				ResNeXt101-FPN & 68.3 & 55.1 & 45.2 & 63.6 & 51.5 & 56.7\\ 
				ResNet50-FPN & 32.8 & 35.5 & 32.7 & 35.7 & 27.6 & 32.9 \\ \hline
				Hourglass104 & 66.4 & 42.1 & 49.7 & 58.7 & 46.9 & 52.76\\
				ResNet101 & 7.4 & 35.8 & 20.5 & 33.6 & 21.7 & 23.8\\
				ResNet18 & 9.6 & 29.5 & 26.3 & 27.9 & 22.1 & 23.1\\ \hline
				$D0$ & 26.9 & 47.0 & 40.5 & 40.3 & 36.8 & 38.3
				
		\end{tabular}}

	\caption{Results on different angle-domains. For example, ResNeXt's AP$_{50}$ performance in medium-right (MR) angles (57-73$^\circ$) is 63.6 AP$_{50}$.}
	\label{table:domain_aps_angle}
	\vspace*{-3mm}
\end{table}

%%%%%
%------------------------------------------------------------------------
\subsection{Single-Object Tracking}

\begin{table*}
	\begin{center}		
		\begin{tabular}{c|ccccccccccccccc}
			Model & MOTA & IDF1 & MOTP & MT & ML & FP & FN & Recall & Prcn & ID Sw. & Frag \\
			\hline
			% FairMOT \cite{zhang2020fairmot} & 45.00 & 45.30 & \textcolor{red}{NaN} & 17 & 20 & 1,975 & 12,324 & 54.10 & 88.00 & 475 & 1,985 \\
			FairMOT-D34 \cite{zhang2020fairmot} & 39.0 & 44.8 & 23.6 & 17 & 17 & 3,604 & 9,445 & 57.2 & 77.8 & 307 & 1,687 \\
			FairMOT-R34 \cite{zhang2020fairmot} & 15.2 & 27.6 & 33.7 & 6 & 37 & 2,502 & 12,592 & 30.1 & 68.4 & 181 & 807 \\
			Tracktor++ \cite{tracktor_2019_ICCV} & 55.0 & 69.6 & 25.6 & 62 & 4 & 7,271 & 3,550 & 85.5 & 74.2 & 165 & 347
		\end{tabular}
	\end{center}
	\caption{Multi-Object Tracking evaluation results for the \textbf{Swimmer} task.}

	\begin{center}		
		\begin{tabular}{c|ccccccccccccccc}
			Model & MOTA & IDF1 & MOTP & MT & ML & FP & FN & Recall & Prcn & ID Sw. & Frag \\
			\hline
			FairMOT-D34 \cite{zhang2020fairmot} & 36.5 & 43.8 & 20.9 & 28 & 49 & 3,788 & 20,867 & 47.2 & 83.1 & 447 & 1,599 \\
			FairMOT-R34 \cite{zhang2020fairmot} & 30.5 & 40.8 & 27.3 & 29 & 127 & 4,401 & 28,999 & 40.2 & 81.6 & 285 & 1,588 \\
            Tracktor++ \cite{tracktor_2019_ICCV} & 71.9 & 80.5 & 20.1 & 123 & 5 & 7,741 & 5,496 & 88.5 & 84.5 & 192 & 438
		\end{tabular}
	\end{center}
	\caption{Multi-Object Tracking evaluation results for the \textbf{All-Objects-In-Water} task.}
	\label{table:mot_results}
	\vspace{-0.4cm}

\end{table*}

Like VisDrone \cite{zhu2020vision}, we provide the success and precision curves for single-object tracking and compare models based on a single number, the success score. As comparison trackers, we choose the DiMP family (DiMP50, DiMP18, PrDiMP50, PrDiMP18) \cite{bhat2019learning,danelljan2020probabilistic} and Atom \cite{danelljan2019atom} because they were the foundation of many of the submitted trackers to the last VisDrone workshop \cite{fan2020visdrone}.

%We take pre-trained versions of these trackers and evaluate them on the testing set.

\begin{figure}
	
	\centering
	\includegraphics[scale=0.16]{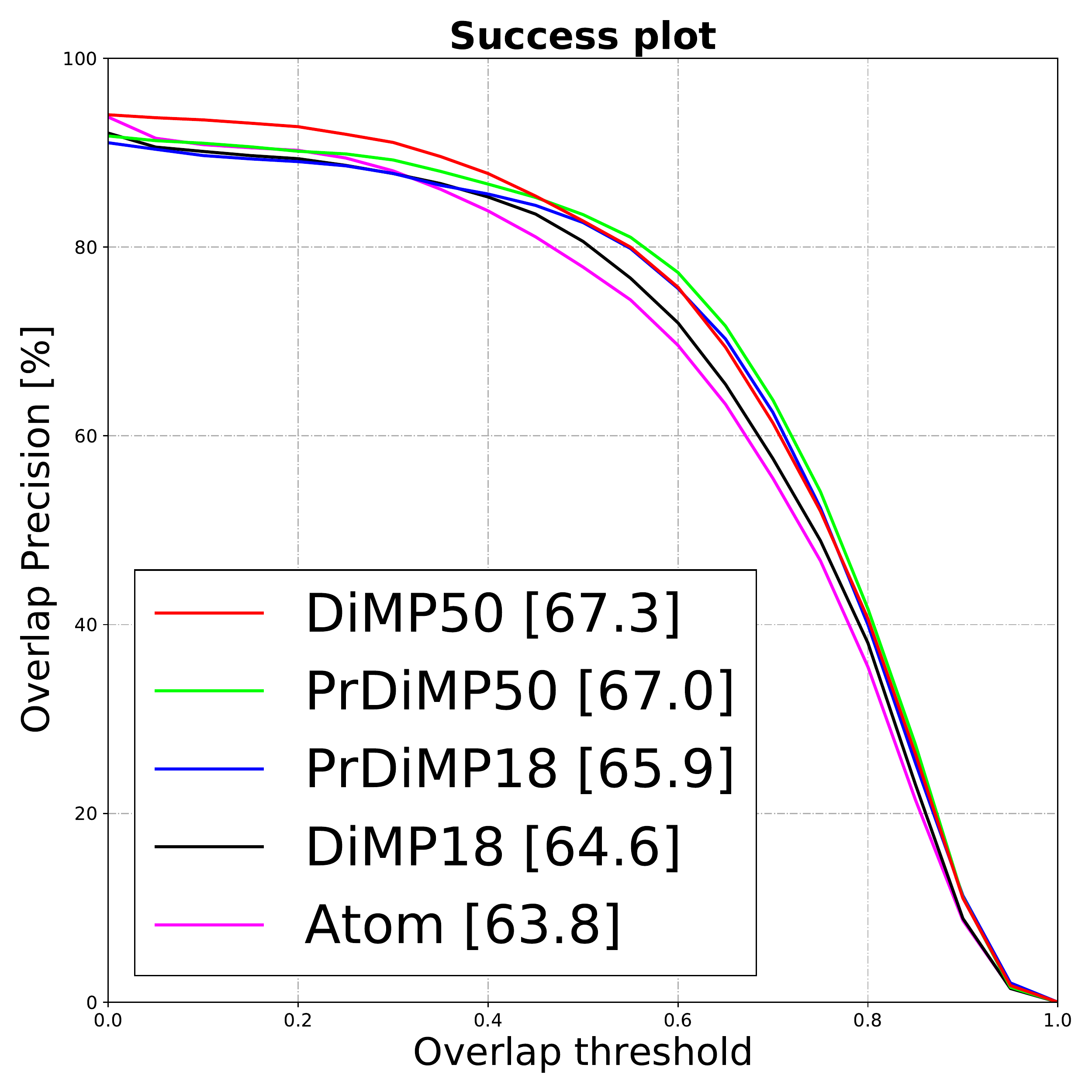}	
	\includegraphics[scale=0.16]{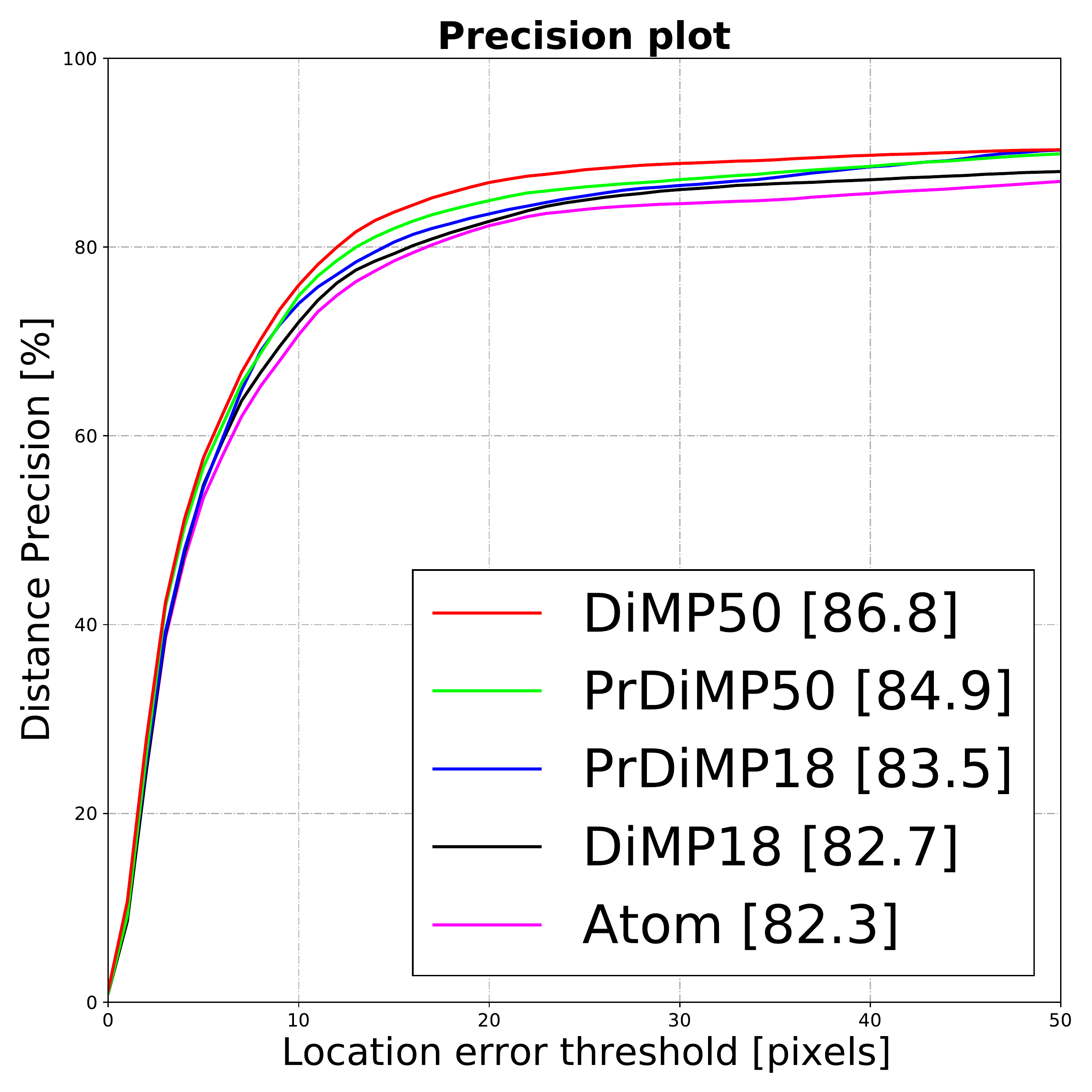}	
	
	\caption{Success and precision plots for single-object tracking task (best viewed in color).}
	\label{fig:sot_results}
	\vspace*{-0.4cm}
\end{figure}

Figure \ref{fig:sot_results} shows that the PrDiMP- and DiMP-family expectedly outperform the older Atom tracker in both, success and precision. Surprisingly, PrDiMP50 slightly trails the accuracy of its predecessor DiMP50. Furthermore, all trackers' performances on SeaDronesSee are similar or worse than on UAV123 (\eg Atom with 65.0 success) \cite{bhat2019learning,danelljan2020probabilistic,danelljan2019atom}, for which they were heavily optimized. We argue that in SeaDronesSee there is still room for improvement, especially considering that the clips feature precise meta information that may be helpful for tracking. Furthermore, in our experiments, the faster trackers DiMP18 and Atom run at approximately 27.1 fps on an Nvidia RTX 2080 Ti. However, we note that they are not capable of running in real-time on embedded hardware, a use-case especially important for UAV-based SAR missions.

\begin{table}

		\resizebox{240pt}{!}{%
			\begin{tabular}{c|ccccc|c}
								Model &  L & LM & M & MH & H & AP$_{50}^{\text{avg}}$ \\ \hline

				\hline
				F. ResNet-50-FPN &\bf 32.8 & 29.8 & 23.5 & 40.5 &\bf 48.9 & 35.1\\ 
		        5$\times$Altitude@3\cite{kiefer2021leveraging} &\bf 32.8 &\bf 29.9 &\bf 26.2 &\bf 41.5 &\bf 48.9 &\bf 35.9 \\

		\end{tabular}}

			\resizebox{240pt}{!}{%
			\begin{tabular}{c|ccccc|cc}
								Model &  A & AM & M & MR & R & AP$_{50}^{\text{avg}}$ \\ \hline

				\hline
				F. ResNet-50-FPN & 32.8 &\bf 35.5 & 32.7 &\bf 35.7 & 27.6 & 32.9\\ 
		        5$\times$Angle@3\cite{kiefer2021leveraging}& \bf 42.0 &\bf 35.5 &\bf 39.3 &\bf 35.7 &\bf 27.7 &\bf 36.0   \\

		\end{tabular}}

	\caption{Results on different altitude- and angle-domains.}% The domain-aware method \cite{kiefer2021leveraging} outperforms the domain-agnostic method.}
	
	\label{table:domain_aps_leveraging}
	\vspace{-0.4cm}
\end{table}

%------------------------------------------------------------------------

\subsection{Multi-Object Tracking}

We use a similar evaluation  protocol as the MOT benchmark \cite{milan2016mot16}. That is, we report results for Multiple Object Tracking Accuracy (MOTA), Identification F1 Score (IDF1), Multiple Object Tracking Precision (MOTP), number of false positives (FP), number of false negatives (FN), recall (R), precision (P), ID switches (ID sw.), fragmentation occurrences (Frag). We refer the reader to \cite{ristani2016performance} or the appendix for a thorough description of the metrics. \\
We train and evaluate FairMOT \cite{zhang2020fairmot}, a popular tracker, which is the base of many trackers submitted to the challenge \cite{fan2020visdrone2}. FairMOT-D34 employs a DLA34 \cite{yu2018deep} as its backbone while FairMOT-R34 makes use of a ResNet34. Another SOTA tracker is Tracktor++ \cite{tracktor_2019_ICCV}, which we also use for our experiments. It performed well on the MOT20 \cite{Dendorfer} challenge and is conceptually simple. \\
Surprisingly, Tracktor++ was better than FairMOT in both tasks. One reason for this may be the used detector. Tracktor++ utilizes a Faster-R-CNN with a ResNet50 backbone. In contrast, FairMOT is using a CenterNet with a DLA34 and a ResNet34 backbone, respectively.

\iffalse
\begin{figure}
	
	\centering
	% left, bottom, right, top
	\includegraphics[scale=0.135,trim=0 600 500 0,clip]{2760_pred.png}

	\caption{Example predictions. Unseen wave patterns result in false positive detections: The detector falsely detects swimmers (light green bounding boxes) in the right part of the image.}
	\label{fig:false_positives}
	%
\end{figure}
\fi

%------------------------------------------------------------------------

\subsection{Meta-Data-Aware Object Detector}

Developing meta-data-aware object detectors is difficult since there are no large-scale data sets to evaluate their performances. However, some works provide promising preliminary results using this metadata \cite{wu2019delving,messmer2021gaining,kiefer2021leveraging}. We provide an initial baseline from \cite{kiefer2021leveraging} incorporating the meta data. We evaluate the performances of 5$\times$Altitude@3- and 5$\times$Angle@3-experts, which are constructed on top of a Faster R-CNN with ResNet-50-FPN, respectively. Essentially, these experts make use of meta-data by allowing the features to adapt to their responsible specific environmental domains.

As Table \ref{table:domain_aps_leveraging} shows, meta data can enhance the accuracy of an object detector considerably. For example, 5$\times$Angle@3 outperforms its ResNet-50-FPN baseline by $3.1$ AP$^{\text{avg}}_{50}$ while running at the same inference speed. The improvements are especially significant for underrepresented domains, such as $+9.2$ and $+6.4$ AP$^{\text{avg}}_{50}$ for the acute angle (A) and the medium angle (M), respectively, which are underrepresented as can be seen from Fig. \ref{fig:meta_distribution}.

%TODO acute ngle
%was meinste damit?

\section{Conclusions}

This work serves as an introductory benchmark in UAV-based computer vision problems in maritime scenarios. We build the first large scaled-data set for detecting and tracking humans in open water. Furthermore, it is the first large-scaled benchmark providing full environmental information for every frame, offering great opportunities in the so-far restricted area of multi-modal object detection and tracking. 
We offer three challenges, object detection, single-object tracking, and multi-object tracking by providing an evaluation server. We hope that the development of meta-data-aware object detectors and trackers can be accelerated by means of this benchmark. 
Moreover, we provide multi-spectral imagery for detecting humans in open water. These images are very promising in maritime scenarios, having the ability to capture wavelengths, which set apart objects from the water background.

%We note, however, that the data can be even more variable. Specifically, we want to emphasize that footage at night or during rain is of great importance in SAR missions. Furthermore, the variance in life jackets, different water types, and subjects with different skin and clothes colors can be improved. We hope that the work at hand attracts more attention to UAV-based computer vision problems in maritime SAR scenarios.

\section*{Acknowledgment}
We would like to thank Sebastian Koch, Hannes Leier and Aydeniz Soezbilir, without whose contribution this work would not have been possible.\\
This work has been supported by the German Ministry for Economic
Affairs and Energy, Project Avalon, FKZ: 03SX481B.

\newpage

{\small
\bibliographystyle{ieee_fullname}
\bibliography{egbib}
}

\end{document}